\PassOptionsToPackage{table}{xcolor}
\PassOptionsToPackage{numbers,compress}{natbib}

\documentclass{article} 
\usepackage{iclr2027_conference,times}

\usepackage{amsmath,amsfonts,bm}

\def\eqref#1{equation~\ref{#1}}

\def\1{\bm{1}}

\DeclareMathAlphabet{\mathsfit}{\encodingdefault}{\sfdefault}{m}{sl}
\SetMathAlphabet{\mathsfit}{bold}{\encodingdefault}{\sfdefault}{bx}{n}

\usepackage{hyperref}
\usepackage{url}
\usepackage{graphicx}
\usepackage{amssymb}
\usepackage{booktabs}
\usepackage{multirow}
\usepackage[table]{xcolor}
\usepackage{subcaption}
\setcitestyle{numbers,square,comma}

\definecolor{darkgreen}{RGB}{0,100,0}
\definecolor{darkblue}{RGB}{0,70,140}

\def\ourmethod{Vantage\xspace}
\def\ourmethodnospace{Vantage}
\def\ourapproach{Seek-and-View\xspace}

\usepackage[most]{tcolorbox}
\usepackage{textcomp}
\tcbuselibrary{listings,breakable}
\providecommand{\ph}[1]{\textcolor{blue!60!black}{\textit{$\langle$#1$\rangle$}}}
\providecommand{\promptinset}{0.01\linewidth}
\newtcblisting[auto counter]{promptbox}[3][]{%
  enhanced jigsaw, breakable, listing only, pad at break=2mm,
  grow to left by=-\promptinset, grow to right by=-\promptinset,
  colback=gray!4, colframe=gray!55!black, boxrule=0.5pt, arc=2pt,
  left=4pt, right=4pt, top=2pt, bottom=2pt,
  fonttitle=\bfseries\small, title={Prompt~\thetcbcounter: #2},
  title after break={Prompt~\thetcbcounter\ (continued)},
  label={#3}, #1,
  listing options={basicstyle=\ttfamily\scriptsize, breaklines=true,
    breakatwhitespace=false, columns=fullflexible, keepspaces=true,
    upquote=true, literate={"}{{\char34}}1, showstringspaces=false, escapeinside={(*}{*)},
    postbreak=\mbox{\textcolor{gray}{$\hookrightarrow$}\space}}}

\RequirePackage{xspace}
\RequirePackage{pifont}
\makeatletter
\DeclareRobustCommand\onedot{\futurelet\@let@token\@onedot}
\def\@onedot{\ifx\@let@token.\else.\null\fi\xspace}

\def\eg{\emph{e.g}\onedot} 

\def\ie{\emph{i.e}\onedot}

\makeatother
\usepackage[dvipsnames]{xcolor}
\hypersetup{
    colorlinks=true,
    urlcolor=Blue,
    citecolor=Blue
}

\title{Seek-and-View Reasoning for Multi-View \\ Spatial Understanding}

\author{
\makebox[\textwidth][s]{%
Qixiang Chen$^{1}$\hfill
Cheng Zhang$^{1}$\hfill
Fucai Ke$^{1}$\hfill
Chi-Wing Fu$^{2}$\hfill
Jianfei Cai$^{1}$\hfill
Jingwen Ye$^{1}$%
} \\
$^{1}$Monash University \\
$^{2}$The Chinese University of Hong Kong
}

\iclrfinalcopy 
\begin{document}

\maketitle
\lhead{Preprint}

\begin{abstract}
    Existing approaches to multi-view spatial reasoning operate largely on sparse input views.
    Vision-language models (VLMs) are thus restricted to understand a scene and infer spatial relations within these fixed views, leading to fragile cross-view alignment and geometry-to-language bottleneck.
    To address these issues, we formulate a novel \emph{\ourapproach} reasoning approach to find implicit cross-view spatial evidence by locating a question-relevant view to support the spatial reasoning.
    To realize this approach, we propose \ourmethod, a training-free model-agnostic reasoning framework that pairs a VLM with a 3D foundation model: 
    a viewpoint-grounded reasoning stage for question analysis and view planning, followed by a geometry-grounded evidence augmentation stage to effectively synthesize and incorporate visual evidence into the final reasoning.
    Comprehensive experiments on six VLMs demonstrate consistent improvements on five benchmarks without fine-tuning.
    Overall, by revealing spatial evidence through view-grounded reasoning, \ourmethod can largely reduce reliance on language-based cross-view alignment and improve multi-view spatial understanding.
    Our code is available at \href{https://github.com/q1xiangchen/Vantage}{https://github.com/q1xiangchen/Vantage}.
\end{abstract}
\section{Introduction}
Multi-view spatial understanding~\citep{gholami2026spatial,zhang2026flatland,wang2026mindcube,ke2026view2space} requires models to capture the structure of a scene from partial observations that are fixed and sparse.
Accurate cross-view alignment is therefore essential.
Especially, spatial questions cannot be answered simply from individual views  and instead require associating evidence from different views.
Despite rapid progress in vision-language models (VLMs), their spatial reasoning over multiple views remains limited and prone to hallucination~\citep{fu2024blink}.

\begin{figure}[ht]
    \centering
    \includegraphics[width=\linewidth]{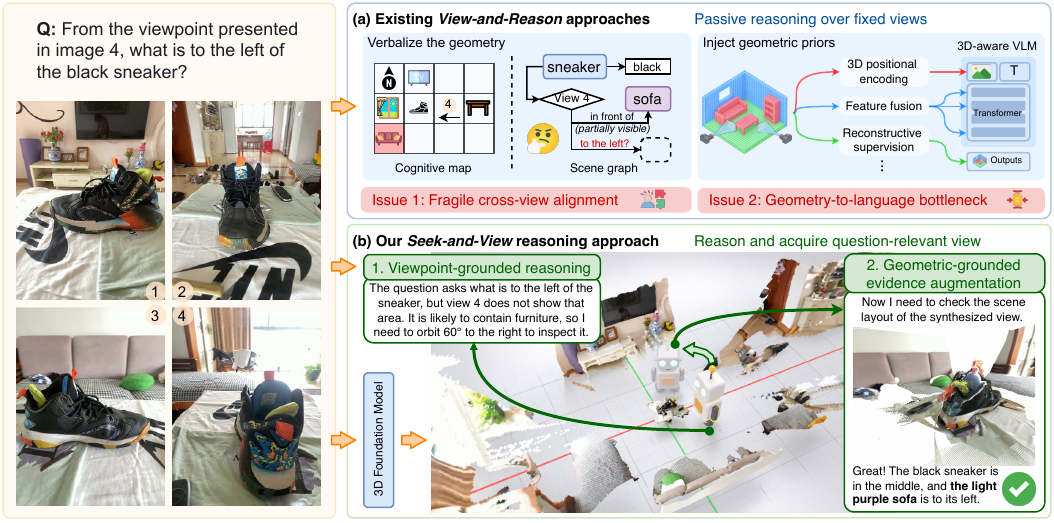}
    \caption{
    (a) Existing approaches reason over fixed views, suffering from fragile cross-view alignment and the geometry-to-language bottleneck.
    (b) Our novel \ourapproach reasoning paradigm seeks a question-relevant view (see bottom right) to make the spatial evidence directly observable.
    }
    \label{fig:teaser}
\end{figure}

Recent VLMs therefore seek to strengthen geometric reasoning capabilities by verbalizing the geometry during reasoning~\citep{ke2026view2space,jia2026omnispatial,li2026omniview} or by injecting geometric priors into the model~\citep{zheng2026learning,zhang2026make,Hu_2026_CVPR}, both aimed at improving the model's awareness of the underlying scene structure (see the illustrations in Figure~\ref{fig:teaser}(a)).
While improving over shallow visual recognition, both approaches rely on the VLM to align views and perform the reasoning explicitly in language, leading to the following two issues. 
First is \emph{fragile cross-view alignment}. Strong semantic and contextual reasoning ability does not inherently enforce structured and globally consistent geometric constraints required across views, leaving the alignment implicit and error-prone.
Second is \emph{the geometry-to-language bottleneck}. Upon compressing the spatial geometry into discrete semantic descriptions, the reasoning would potentially overlook fine-grained spatial information~\citep{chen2025openview,wu2026reinforcing,chen2026geometrically}.
More fundamentally, both approaches remain within a \textbf{View-and-Reason} paradigm, in which the reasoning is fixed on the given observations.
Humans, in contrast, can conceptualize the three-dimensional structure of a scene and seek a view to reveal the visible information for resolving the spatial relation.
Such a visual thinking process leads us to ask: \emph{Can a model decide where to look, then reason better from that view?}

To explore this idea,
we formulate a novel \textbf{\ourapproach} paradigm, in which a model first seeks a viewpoint that consolidates evidence from different views into an observation for supporting the spatial reasoning.
Our new paradigm alleviates the burden on semantic textual reasoning for global scene understanding through question-relevant, evidence-seeking, and reasoning-oriented visual evidence acquisition.
Recent studies explore a think-with-image paradigm for visual question answering (VQA), where the model reasons with additional visual cues obtained by zooming, drawing auxiliary lines, or estimating depth~\citep{gupta2023visual,suris2023vipergpt,openai2025thinkingwithimages,qin2025chain,wu2026reinforcing,hu2026omni}.
Yet, the underlying visual thoughts are still an in-view understanding confined to the fixed observations.
Beyond fixed observations, we aim to realize a \ourapproach paradigm, in which the challenges are 
(i) to locate a question-relevant viewpoint, and 
(ii) to effectively contextualize the acquired visual evidence for subsequent reasoning.

To address these challenges, we propose \textbf{\ourmethod}, a two-stage training-free and model-agnostic framework that instantiates the \ourapproach paradigm by incorporating a VLM with a 3D foundation model (3DFM).
In the first \emph{viewpoint-grounded reasoning} stage, the VLM analyzes the question to identify the necessary observation(s) and plans a viewpoint to acquire the intended evidence.
In the second \emph{geometry-grounded evidence augmentation} stage, the 3DFM produces a new view of the scene and takes it with the given question and proper guidance to support the final VLM reasoning.
In this way, the 3DFM externalizes the cross-view alignment by reconstructing a consistent scene representation, meanwhile the VLM plans the viewpoint and continues reasoning with the additional visual evidence acquired from the new view; see Figure~\ref{fig:teaser}(b).

Our evaluations focus on multi-view spatial VQA tasks whose answers are determined by the geometric relations within static real-world scenes.
Across five spatial reasoning benchmarks and six different VLMs, \ourmethod is able to consistently boost the performance of all the six VLMs without any fine-tuning, with a mean accuracy gain of 6.7\%.
These results highlight that the difficulty in current multi-view spatial understanding stems not only from limited geometric reasoning capability but also from passive reasoning on fixed input observations.
Further, we analyze failure cases and reveal the other limitations of existing VLMs.
Our main contributions are threefold:
\begin{itemize}
    \item We introduce a novel \ourapproach reasoning paradigm for multi-view spatial understanding, in which we propose to seek cross-view spatial evidence through a question-relevant viewpoint rather than reasoning solely on fixed input observations.
    \item We propose \ourmethod, a two-stage training-free framework to realize \ourapproach with a viewpoint-grounded reasoning and a geometry-grounded evidence augmentation, explicitly coupling VLM reasoning with a 3D foundation model.
    \item Through extensive experiments across six VLMs and five benchmarks, we demonstrate the effectiveness of \ourapproach, revealing its strength to boost the VLM performance and also the limitations of existing VLMs under this reasoning paradigm.
\end{itemize}
\section{Related Work}
\paragraph{Multi-View Spatial Reasoning}
Existing research has enhanced VLMs for multi-view spatial reasoning largely in two directions.
The first verbalizes scene geometry into textual or symbolic structures, such as cognitive maps, scene graphs, or bounding boxes to support downstream reasoning~\citep{lee2025perspective,ke2026view2space,wang2026mindcube,li2026omniview,tao2026viewfusion,cai2026scaling,cheng2026grounded,choi2026dense}.
This approach, however, demands labor-intensive per-scene annotations and subsequent supervised fine-tuning.
The second direction injects geometric priors directly into the model~\citep{wang2025ross3d,zheng2026learning,janjua2026don,Hu_2026_CVPR,zhang2026spatialstack,zhang2026make},~\eg, by fusing features from geometric encoders, redesigning positional encoding with 3D coordinates, or introducing reconstructive training objectives.
Despite these advances, the visual observations are typically fixed, leaving the model to reason over a predetermined set of views.
In contrast, our \ourapproach reasoning paradigm adapts the observation itself by acquiring a task-specific view tailored to the question, while keeping the core VLM training-free.

\paragraph{Thinking with images paradigm} 
Thinking with images~\citep{openai2025thinkingwithimages} is an emerging paradigm that externalizes intermediate reasoning into visual operations such as sketching, marking, zooming, and depth estimation for mathematical geometry and 2D spatial tasks~\citep{gupta2023visual,suris2023vipergpt,hu2024visual,qin2025chain,zhang2026policy,yang2026machine,ke2025explain,wu2026reinforcing,Ke_2025_dwim}.
Related agentic approaches further treat the VLM as a tool-calling agent, augmenting reasoning via various external tools~\citep{chen2026geometrically,chen2026enhancing,cho2026spatialclaw,wang2026vagen}.

Spatial reasoning can leverage a world model to generate unseen observations for iterative exploration~\citep{yang2026mindjourney,cao2026spatialdreamer,zhang2026world2vlm,zhu2026thinking, liu2026abstract,zhang2026thinking,hu2026omni}.
This approach, however, incurs substantial inference cost, while requiring dedicated fine-tuning to maintain cross-view consistency.
Other approaches~\citep{yang2026and,zhang2026think3d,jin2026spazer} construct stitched panoramic views or top-down scene representations to try to preserve the scene consistency.
Yet, these representations are typically generated as generic global views, rather than tailored to expose the spatial evidence needed for a specific question.
Our approach instead performs a question-relevant view seeking, allowing the geometric backend to acquire the desired visual evidence to support the subsequent spatial reasoning.

\paragraph{3D Foundation Models for View Synthesis}
Recent 3D foundation models (3DFMs)~\citep{wang2025vggt,lin2025depth,wang2025pi,wang2026vggt} have made feed-forward recovery of scene geometry increasingly practical.
G3T~\citep{kani2026g3t} further introduces gravity-aligned point maps, providing an upright scene coordinate frame.
By reconstructing a consistent scene representation and estimating camera poses, 3DFMs externalize the cross-view alignment and allow viewpoint planning and view synthesis of the scene.
Such a view synthesis differs from conventional novel view synthesis, which aims to faithfully render a scene from a specified pose.
Our setting instead seeks a viewpoint mainly for  exposing the spatial evidence to support the reasoning.
Hence, the synthesized view need not be photorealistic.
Its core value is to preserve the underlying scene and reveal necessary visual evidence needed for answering the given question.
\section{Method}
\begin{figure}[t]
    \centering
    \includegraphics[width=\textwidth]{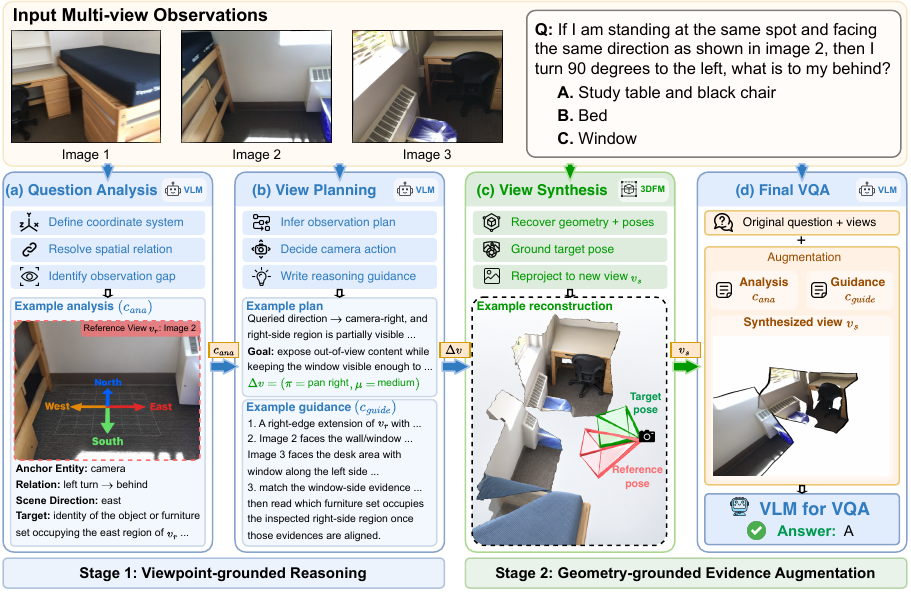}
    \caption{\textbf{Overview of \ourmethod pipeline.} Stage 1: Viewpoint-grounded reasoning employs an VLM (in blue) to (a) produce an analysis on the input question (top-right), select reference view $v_r$, identify queried spatial relation, then (b) plan a question-relevant camera action $\Delta v$ on reference view $v_r$ and generate reasoning guidance for supporting view interpretation. Stage 2: (c) Geometry-grounded evidence augmentation synthesizes the planned view $v_s$ from the 3DFM reconstruction (in green) and provides it together with the original images, viewpoint analysis $c_{\mathrm{ana}}$, and reasoning guidance $c_{\mathrm{guide}}$ to support (d) the final reasoning.
    }
    \label{fig:pipeline}
\end{figure}

\ourmethod is a two-stage pipeline, as shown in Figure~\ref{fig:pipeline}.
Its name reflects the key idea,~\ie, the model seeks a vantage viewpoint from which the spatial relation relevant to answering the given question can become visible to support the reasoning.
In this section, we first overview the \ourapproach reasoning paradigm in Section~\ref{sec:formulation}, followed by the method details in Sections~\ref{sec:view_planning} and~\ref{sec:view_acquisition}.

\subsection{Overview of \ourapproach Reasoning Paradigm}
\label{sec:formulation}

Given a spatial question $q$ and a set of sparse, unposed input views $\mathcal{V}=\{v_i\}_{i=1}^{N}$, we consider visual question answering (VQA), in which the observed input views $\mathcal{V}$ collectively provide the scene information for answering question $q$. 
However, the relevant spatial relation for answering question $q$ may not be directly observable solely from any individual view in $\mathcal{V}$.

\paragraph{Seeking.} \
Directly using a VLM to predict the camera pose of the target view is unreliable~\citep{zhang2026thinking}.
Yet, introducing a metric 3D coordinate system could further escalate the uncertainty in scale, orientation, and alignment.
We thus formulate view planning as a relative camera action grounded in one of the observed views to more reliably control the viewpoint.
A reasoner $\mathcal{H}$ analyzes the question $q$ and available observations $\mathcal{V}$ to select a reference view $v_r$ and a relative camera action $\Delta v$, while retaining intermediate reasoning contexts for subsequent answer prediction:
\begin{equation}
    \mathcal{H}(q,\mathcal{V})
    \rightarrow
    (v_r, \Delta v, c_{\mathrm{ana}}, c_{\mathrm{guide}}),
\end{equation}
where $c_{\mathrm{ana}}$ captures the question-oriented analysis of the current observations, and $c_{\mathrm{guide}}$ contains the planned view adjustment and its underlying rationale.

\paragraph{Viewing.} \
Before answering the input question $q$, we obtain the view through a geometry-grounded synthesis: a 3D foundation model reconstructs the scene and its camera poses from the unposed input views $\mathcal{V}$, and the planned action $\Delta v$ is applied to the reference view $v_r$ to reproject the reconstruction to produce the requested view $v_s$.
The synthesized view $v_s$ is then incorporated into the original VQA $(q, \mathcal{V})$, together with the intermediate reasoning contexts $(c_{\mathrm{ana}}, c_{\mathrm{guide}})$, to support the subsequent spatial reasoning.
We detail the reconstruction and reprojection in Section~\ref{sec:view_acquisition}.

\subsection{Viewpoint-Grounded Reasoning}
\label{sec:view_planning}
Instead of unconstrained reasoning, we structure view planning around question analysis and action planning in viewpoint-aware perspective.
In this way, we can resolve spatial relations not directly observable from any individual view and determine how to adjust the viewpoint to reveal the spatial relations.
However, grounding such semantic reasoning into an executable camera action remains challenging. 
We therefore decompose viewpoint-grounded reasoning into the following two steps:

\noindent\textbf{Question analysis:}
\textit{Establishing a semantic directional frame from the reference view.}
A natural starting point for multi-view reasoning is to find a view whose scene contents are most relevant to the question. 
Here, we employ a VLM to first select a reference view $v_r$ from the input views $\mathcal{V}$ conditioned on the question $q$.
It then then identifies the anchor entity, namely the camera viewpoint, object, or region in the reference view that constitutes one side of the queried spatial relation.
Directional terms such as left, right, front, and behind may refer to different frames of reference while sharing the same linguistic expressions~\citep{yang2026mmsi,wang2026mindcube}.
This creates ambiguity in how a directional relation should be spatially grounded.
To reduce this ambiguity, the VLM establishes a semantic directional frame grounded in reference view $v_r$. By default, the forward direction of $v_r$ is treated as north, unless an explicit directional convention is specified by the question. This semantic frame is maintained throughout the reasoning process to provide a consistent interpretation of directional relations without requiring explicit metric 3D coordinates.

\noindent\textit{Resolving Spatial Relations into Observation Needs.}
The VLM identifies the queried entity or region and the spatial relation with respect to the anchor. It then resolves this relation within the established semantic directional frame, translating object-relative or camera-relative directions into a unified scene direction; see Figure~\ref{fig:pipeline}~(Stage 1).
From the reference view $v_r$, the model examines the currently visible evidence and determines which spatial relation remains visible or verifiable.
Together, these elements form the question analysis context $c_{\mathrm{ana}}$, which summarizes the problem diagnosis and provides the reasoning basis for subsequent viewpoint planning.

\noindent\textbf{View Planning:}
\textit{From Observation Need to Camera Action.}
The observation need identified above is then translated into a camera action. However, it is expressed semantically, whereas acquiring a new view requires a camera pose defined in the reconstructed 3D space.
We therefore provide the VLM with a predefined action set of 17 camera motions spanning rotation, translation, and hybrid transformations, each parameterized by three coarse magnitude levels: small, medium, and large. For example, a rightward rotation corresponds to $30^\circ$, $60^\circ$, and $90^\circ$, respectively. Conditioned on question $q$, input views $\mathcal{V}$, question analysis context $c_{\mathrm{ana}}$, and an action-planning prompt that specifies this action set, the VLM predicts a camera adjustment relative to the reference view $v_r$:
\begin{equation}
    \Delta v = (\pi, \mu),
\end{equation}
where $\pi$ denotes the selected camera action and $\mu$ its magnitude level.
After the 3DFM estimates the pose of the reference view, the geometric backend maps $\Delta v$ to the corresponding numerical pose transformation in the reconstructed coordinate frame. This structured interface allows the VLM to reason about how to adjust the viewpoint without directly predicting the metric camera pose, while leaving precise geometric grounding to the reconstruction backend.
Section~\ref{sec:ablation} shows the benefit of this structured action interface and the Appendix provides the action set and planning prompt.

\noindent\textit{Preserving the Reasoning Context.}
The synthesized view alone does not convey why the viewpoint was selected or which spatial cues should be examined. To explicitly convey this planning intent~\citep{wang2026vagen}, the VLM also generates the reasoning guidance context $c_{\mathrm{guide}}$ within the same view planning call.
The guidance describes the expected spatial configuration, identifies relevant visual anchors in the input views, and specifies the spatial cues to inspect or combine in the synthesized view. Since it is generated before view synthesis, it records the intended use of the new observation without assuming that the expected evidence is present.
The resulting $c_{\mathrm{guide}}$ is passed to the final VQA stage as auxiliary context, preserving the connection between the planned observation and the original visual evidence.

\subsection{Geometry-Grounded Evidence Augmentation}
\label{sec:view_acquisition}
In this stage, we ground the planned viewpoint in the reconstructed 3D scene to synthesize the view. We then use this geometry-grounded observation with proper guidance from the previous stage to augment the original 
VQA for continuous reasoning through the following two steps.

\noindent\textbf{View Synthesis:}
\textit{Realizing the Requested Observation.}
It is important that the planned semantic action preserves its intended orientation with respect to the ground plane. For example, the action in Figure~\ref{fig:pipeline} requires the camera to pan $60^\circ$ to the right while remaining horizontally aligned with the ground plane. 
This motivates gravity-aligning the reconstructed scene to provide a canonical world frame for converting semantic actions into geometric camera transformations.
The 3DFM $\mathcal{G}$ first recovers the scene geometry and camera poses from the unposed input views $\mathcal{V}$:
\begin{equation}
    \mathcal{G}(\mathcal{V}) = (\mathcal{Z}, \mathcal{P}),
\end{equation}
where $\mathcal{Z}$ denotes the reconstructed scene as a point cloud and $\mathcal{P}=\{p_i\}_{i=1}^{N}$ the estimated camera poses.
The planned action $\Delta v$ is then deterministically mapped to a 6-DoF camera transformation relative to the reference view pose $p_r$ in this gravity-aligned reconstruction, with the scene geometry providing the scale required to instantiate its coarse magnitude.
Reprojecting the point cloud from the resulting target pose gives the synthesized view:
\begin{equation}
v_s = \mathcal{R}\left(\mathcal{Z}, p_r \oplus \Delta v \right),
\end{equation}
where $\mathcal{R}$ denotes the reprojection operator and $\oplus$ the relative camera transformation induced by $\Delta v$.
Since $v_s$ only rearranges pixels observed in the input views, scene surfaces not covered by any input view remain empty, so the synthesized view supplements rather than replaces the input observations.
More implementation details of view synthesis are provided in the Appendix.

\noindent\textbf{Final VQA:}
\textit{Answering with Contextual Augmentation.}
For final answer prediction, we augment the original inputs with the synthesized view $v_s$ and the reasoning contexts produced during question analysis and view planning. Specifically, $c_{\mathrm{ana}}$ preserves the resolved directional frame and observation need, while $c_{\mathrm{guide}}$ conveys how the synthesized view should be interpreted with respect to the relevant anchors and spatial cues.
The VLM $\mathcal{F}$ then answers the question with the synthesized view and the reasoning contexts:
\begin{equation}
\hat{y} = \mathcal{F}\left(q, \mathcal{V}, v_s, c_{\mathrm{ana}}, c_{\mathrm{guide}}\right),
\label{eq:final_vqa}
\end{equation}
where $\hat{y}$ is the predicted answer.
The original input views remain available throughout this stage, since the synthesized view may be incomplete, allowing the model to combine evidence from both the observed views and the synthesized observation. These three augmentation components, namely the synthesized view, question analysis, and reasoning guidance, are evaluated separately in our ablation study in Section~\ref{sec:ablation}.
\section{Experiments}

\subsection{Setup}

\noindent\textbf{Implementation Details.}
We evaluate \ourmethod on six VLMs for a wide range of model sizes,~\ie, GPT-5.4~\citep{openai2026gpt54}, Gemma-4-31B~\citep{team2026gemma}, Qwen3.6-27B~\citep{qwen36_27b}, InternVL3-8B~\citep{zhu2025internvl3}, Qwen3-VL-8B, and Qwen3-VL-4B~\citep{yang2025qwen3}. 
For the open-source models, we run inference using vLLM~\citep{kwon2023efficient} under the official configurations with two NVIDIA H200 GPUs.
We employ G3T~\citep{kani2026g3t} as the default 3D foundation model (3DFM) for gravity-aligned scene reconstruction and camera pose estimation, and deploy it on the same two GPUs.

\noindent\textbf{Benchmarks.}
We primarily evaluate on multi-view spatial VQA in static scenes, excluding tasks that involve temporal dynamics or simulated environments.
Our main evaluation covers MindCube-tiny~\citep{wang2026mindcube}, the Positional Relationship and Attribute subsets of MMSI-Bench~\citep{yang2026mmsi}, and the Multi-view Reasoning subset of BLINK~\citep{fu2024blink}. 
To evaluate the broader applicability of the \ourapproach paradigm,
we additionally include the Perspective Taking subset of OmniSpatial~\citep{jia2026omnispatial} for single-view spatial understanding, and the Dynamic Rotation and Dynamic Translation subsets of SPINBench~\citep{zhang2026spinbench}, where objects or subjects rotate or translate while the background scene remains static.

We compare with the state-of-the-art methods~\citep{cai2026scaling,chen2026geometrically} on three multi-view benchmarks.
Following the evaluation protocols of prior work~\citep{wang2026mindcube,chen2026geometrically}, we report accuracy (\%) for each subset and the average accuracy (\%) across subsets.
Unless otherwise specified, all evaluations use the benchmark subsets described above.
Details on each benchmark are provided in the Appendix.

\subsection{Main Results}
\label{sec:main_results}

\begin{table*}[t]
  \centering
  \small
  \setlength{\tabcolsep}{4pt}
  \caption{\textbf{Main results on five benchmarks.}
  For each base model, the better result is shown in \textbf{bold}. The final two columns report the average accuracy across all five benchmarks and the relative improvement. 
  $^{\dagger}$Results are reported as in their original papers.
  }
  \label{tab:main}
  \resizebox{\linewidth}{!}{
  \begin{tabular}{lcccccccc|cccc|cc}
  \toprule
  \multirow{2.5}{*}{\textbf{Model}} & \multicolumn{4}{c}{\textbf{MindCube-tiny}} & \multicolumn{3}{c}{\textbf{MMSI-Bench}} & \multicolumn{1}{c}{\textbf{BLINK}} & \multicolumn{1}{c}{\textbf{OmniSpatial}} & \multicolumn{3}{c}{\textbf{SPINBench}} & \multirow{2.5}{*}{\textbf{Avg.}} & \multirow{2.5}{*}{\textbf{\% Gain}} \\
  \cmidrule(lr){2-5} \cmidrule(lr){6-8} \cmidrule(lr){9-9} \cmidrule(lr){10-10} \cmidrule(lr){11-13}
  & Rot. & Ard. & Amg. & All & PR. & Attr. & All & MV & Pers. & D.Rot & D.Tr & All & & \\
  \midrule
  \textit{\textbf{Training-Scaled Spatial Model}} \\
  \quad SenseNova-SI-1.5-InternVL3-8B &  92.5 &  90.8 &  94.5 &  93.24 &  47.1 & 37.7 &  45.2 &  63.9 & 52.0 & 43.6 & 41.7 & 43.0 & 59.5 & - \\
  \textit{\textbf{Model-Scaled Spatial Agent}} \\
  \quad GCA (Qwen3-VL-235B-A22B-Thinking)$^{\dagger}$   & 82.0 & 61.8 & 59.8 & 64.2 &  52.8 &  45.0 & 51.2 & - & 58.6 & - & - & - & - & - \\
  \midrule
  \textit{\textbf{\ourmethod on base models}} \\
  \quad Qwen3-VL-4B & 33.0 & 34.0 & 20.3 & 26.0 & 27.4 & 30.0 & 27.9 & 41.4 & 42.1 & 48.2 & \textbf{64.7} & 53.2 & 38.1 & -- \\
  \quad \cellcolor{gray!6}\hspace{0.6em}+ \ourmethod & \cellcolor{gray!6}\textbf{35.0} & \cellcolor{gray!6}\textbf{46.8} & \cellcolor{gray!6}\textbf{36.8} & \cellcolor{gray!6}\textbf{38.9} & \cellcolor{gray!6}\textbf{34.7} & \cellcolor{gray!6}30.0 & \cellcolor{gray!6}\textbf{33.7} & \cellcolor{gray!6}\textbf{42.1} & \cellcolor{gray!6}\textbf{47.2} & \cellcolor{gray!6}\textbf{49.3} & \cellcolor{gray!6}62.8 & \cellcolor{gray!6}\textbf{53.4} & \cellcolor{gray!6}\textbf{43.1} & \cellcolor{gray!6}$+13.1$ \\
  \midrule
  \quad Qwen3-VL-8B & 36.5 & \textbf{33.2} & 28.8 & 31.3 & \textbf{31.4} & 24.6 & 30.1 & 44.4 & 43.9 & \textbf{44.2} & 73.7 & \textbf{53.2} & 40.6 & -- \\
  \quad \cellcolor{gray!6}\hspace{0.6em}+ \ourmethod & \cellcolor{gray!6}36.5 & \cellcolor{gray!6}28.8 & \cellcolor{gray!6}\textbf{31.2} & \cellcolor{gray!6}\textbf{31.6} & \cellcolor{gray!6}29.9 & \cellcolor{gray!6}\textbf{30.8} & \cellcolor{gray!6}30.1 & \cellcolor{gray!6}\textbf{45.1} & \cellcolor{gray!6}\textbf{45.1} & \cellcolor{gray!6}42.5 & \cellcolor{gray!6}\textbf{75.0} & \cellcolor{gray!6}52.5 & \cellcolor{gray!6}\textbf{40.9} & \cellcolor{gray!6}$+0.7$ \\
  \midrule
  \quad InternVL3-8B & \textbf{35.5} & \textbf{47.2} & 32.8 & 36.8 & 29.3 & 24.6 & 28.4 & 54.1 & \textbf{44.0} & 47.0 & \textbf{53.2} & 48.9 & 42.4 & -- \\
  \quad \cellcolor{gray!6}\hspace{0.6em}+ \ourmethod & \cellcolor{gray!6}31.5 & \cellcolor{gray!6}41.2 & \cellcolor{gray!6}\textbf{42.2} & \cellcolor{gray!6}\textbf{39.9} & \cellcolor{gray!6}\textbf{30.8} & \cellcolor{gray!6}\textbf{27.7} & \cellcolor{gray!6}\textbf{30.2} & \cellcolor{gray!6}\textbf{55.6} & \cellcolor{gray!6}43.9 & \cellcolor{gray!6}\textbf{52.1} & \cellcolor{gray!6}49.4 & \cellcolor{gray!6}\textbf{51.3} & \cellcolor{gray!6}\textbf{44.2} & \cellcolor{gray!6}$+4.2$ \\
  \midrule
  \quad Qwen3.6-27B & 90.5 & \textbf{71.2} & 50.5 & 63.0 & 42.5 & 41.5 & 42.3 & 28.6 & 52.6 & \textbf{85.3} & 98.1 & \textbf{89.2} & 55.1 & -- \\
  \quad \cellcolor{gray!6}\hspace{0.6em}+ \ourmethod & \cellcolor{gray!6}\textbf{93.5} & \cellcolor{gray!6}70.0 & \cellcolor{gray!6}\textbf{60.0} & \cellcolor{gray!6}\textbf{68.8} & \cellcolor{gray!6}\textbf{44.3} & \cellcolor{gray!6}\textbf{46.9} & \cellcolor{gray!6}\textbf{44.8} & \cellcolor{gray!6}\textbf{29.3} & \cellcolor{gray!6}\textbf{64.7} & \cellcolor{gray!6}83.6 & \cellcolor{gray!6}98.1 & \cellcolor{gray!6}88.0 & \cellcolor{gray!6}\textbf{59.1} & \cellcolor{gray!6}$+7.3$ \\
  \midrule
  \quad Gemma-4-31B & 51.0 & \textbf{63.6} & 50.2 & 53.5 & 35.4 & \textbf{42.3} & 36.8 & 36.1 & 51.7 & 59.8 & \textbf{96.8} & 71.1 & 49.8 & -- \\
  \quad \cellcolor{gray!6}\hspace{0.6em}+ \ourmethod & \cellcolor{gray!6}\textbf{82.0} & \cellcolor{gray!6}56.0 & \cellcolor{gray!6}\textbf{57.7} & \cellcolor{gray!6}\textbf{61.9} & \cellcolor{gray!6}\textbf{36.2} & \cellcolor{gray!6}40.8 & \cellcolor{gray!6}\textbf{37.1} & \cellcolor{gray!6}\textbf{44.4} & \cellcolor{gray!6}\textbf{55.1} & \cellcolor{gray!6}\textbf{64.9} & \cellcolor{gray!6}94.2 & \cellcolor{gray!6}\textbf{73.9} & \cellcolor{gray!6}\textbf{54.5} & \cellcolor{gray!6}$+9.4$ \\
  \midrule
  \quad GPT-5.4 & 38.5 & 50.8 & 43.3 & 44.2 & 33.3 & 38.5 & 34.4 & 48.1 & 47.4 & 49.0 & 82.7 & 59.3 & 46.7 & -- \\
  \quad \cellcolor{gray!6}\hspace{0.6em}+ \ourmethod & \cellcolor{gray!6}\textbf{61.5} & \cellcolor{gray!6}\textbf{56.8} & \cellcolor{gray!6}\textbf{50.8} & \cellcolor{gray!6}\textbf{54.3} & \cellcolor{gray!6}\textbf{41.8} & \cellcolor{gray!6}38.5 & \cellcolor{gray!6}\textbf{41.1} & \cellcolor{gray!6}\textbf{54.1} & \cellcolor{gray!6}\textbf{53.7} & \cellcolor{gray!6}\textbf{53.8} & \cellcolor{gray!6}\textbf{86.5} & \cellcolor{gray!6}\textbf{63.9} & \cellcolor{gray!6}\textbf{53.4} & \cellcolor{gray!6}$+14.3$ \\
  \bottomrule
  \end{tabular}
  }
\end{table*}

\noindent\textbf{Consistent Gains Across Benchmarks.}
Table~\ref{tab:main} shows that \ourmethod consistently improves the average accuracy of all six VLMs without additional training.
The gains extend to strong base performance, with Qwen3.6-27B improving from 90.5\% to 93.5\% on the MindCube-tiny Rotation subset.
More substantial gains are observed when cross-view reasoning remains challenging, as Gemma-4-31B improves from 51.0\% to 82.0\% on the same subset.
Beyond the multi-view benchmarks, \ourmethod also improves five of the six models on OmniSpatial, where viewpoint adjustment can expose inferred 3D structures through parallax and provide auxiliary depth cues.
On the dynamic SPINBench subsets, \ourmethod improves four of the six models, suggesting that the sought view helps to provide spatial cues for object-motion reasoning.
Overall, these results demonstrate the consistent benefits of \ourmethod through viewpoint-grounded reasoning and geometry-grounded evidence augmentation to enhance the spatial reasoning.

\noindent\textbf{Comparison with State-of-the-Art Methods.}
We also compare with state-of-the-art methods on three multi-view benchmarks.
These approaches achieve strong performance through either extensive spatial-task training or tool-integrated agents built on substantially larger VLMs.
In contrast, \ourmethod achieves its improvements with a training-free and diagnostic pipeline, reaching comparable or superior performance on several subsets without scaling task-specific training or model capacity for reliable tool calling.
This highlights the potential of \ourapproach as a complementary direction for advancing multi-view spatial reasoning.

\noindent\textbf{Analysis and Planning Are Critical to \ourapproach.}
The gains vary across models and do not increase monotonically with model size.
To study further into this behavior, we compose the two steps in Stage 1,~\ie, question analysis and view planning, using different model sources, while keeping Stage 2 unchanged.
As shown in Table~\ref{tab:xfer_ablation}, the additional gains for the two smaller models depend strongly on the quality of question analysis and view planning, particularly for Qwen3-VL-8B.
Directly replacing both steps with outputs from the larger Qwen3.6-27B does not always yield the best performance.
Instead, stronger results are obtained often when the smaller model performs view planning based on the question analysis by Qwen3.6-27B.
This suggests that the camera action and reasoning guidance should be aligned with the reasoning needs of the model for final VQA.

\begin{table}[t]
  \centering
  \small
  \setlength{\tabcolsep}{11pt}
  \caption{\textbf{Cross-source composition results.}
  \textit{Self} denotes steps executed by the evaluated model, while \textit{Cross} denotes steps replaced by Qwen3.6-27B.
  Bold indicates the highest accuracy within each model group.
  $\Delta$ denotes the accuracy change relative to model + \ourmethodnospace.}
  \label{tab:xfer_ablation}
  \resizebox{\linewidth}{!}{
  \begin{tabular}{l|cc|rrrrrrrr}
  \toprule
  \multirow{2.5}{*}{\textbf{Model}}
    & \multirow{2.5}{*}{\textbf{\shortstack{Question\\Analysis}}}
    & \multirow{2.5}{*}{\textbf{\shortstack{View\\Planning}}}
    & \multicolumn{2}{c}{\textbf{MindCube-tiny}}
    & \multicolumn{2}{c}{\textbf{MMSI-Bench}}
    & \multicolumn{2}{c}{\textbf{BLINK}}
    & \multicolumn{2}{c}{\textbf{OmniSpatial}} \\
  \cmidrule(lr){4-5}
  \cmidrule(lr){6-7}
  \cmidrule(lr){8-9}
  \cmidrule(lr){10-11}
    & & & Acc. & $\Delta$ & Acc. & $\Delta$ & Acc. & $\Delta$ & Acc. & $\Delta$ \\
  \midrule
  \multirow{4}{*}{Qwen3-VL-4B}
    & -- & -- & 26.0 & $-12.9$ & 27.9 & $-5.8$ & 41.4 & $-0.7$ & 42.1 & $-5.1$ \\
    & Self & Self & 38.9 & -- & 33.7 & -- & \textbf{42.1} & -- & 47.2 & -- \\
    & Cross & Self & \textbf{51.0} & $+12.1$ & 35.3 & $+1.6$ & 37.6 & $-4.5$ & \textbf{50.6} & $+3.4$ \\
    & Cross & Cross & 49.9 & $+11.0$ & \textbf{36.2} & $+2.5$ & 36.8 & $-5.3$ & 48.1 & $+0.9$ \\
  \midrule
  \multirow{4}{*}{Qwen3-VL-8B}
    & -- & -- & 31.3 & $-0.3$ & 30.1 & $0.0$ & 44.4 & $-0.7$ & 43.9 & $-1.2$ \\
    & Self & Self & 31.6 & -- & 30.1 & -- & \textbf{45.1} & -- & 45.1 & -- \\
    & Cross & Self & 50.6 & $+19.0$ & \textbf{35.1} & $+5.0$ & 36.1 & $-9.0$ & \textbf{51.3} & $+6.2$ \\
    & Cross & Cross & \textbf{50.8} & $+19.2$ & 34.8 & $+4.7$ & 33.1 & $-12.0$ & 49.0 & $+3.9$ \\
  \midrule
  \multirow{2}{*}{Qwen3.6-27B}
    & -- & -- & 63.0 & $-5.8$ & 42.3 & $-2.5$ & 28.6 & $-0.7$ & 52.6 & $-12.1$ \\
    & Self & Self & \textbf{68.8} & -- & \textbf{44.8} & -- & \textbf{29.3} & -- & \textbf{64.7} & -- \\
  \bottomrule
  \end{tabular}}
\end{table}

\subsection{Ablation Studies}
\label{sec:ablation}
We evaluate the design of \ourmethod through ablations of its question analysis step, action interface, and final VQA context. 
Reconstruction backend comparisons are provided in the Appendix.

\noindent\textbf{Role of the Question Analysis.}
As shown in Table~\ref{tab:per_module_ablation}, we compare the full pipeline with \emph{QA-skip}, which plans the view without question analysis, and \emph{QA-only}, which uses question analysis alone to augment the final VQA.
For Qwen3-VL-4B, removing QA reduces accuracy from 38.9\% to 30.0\% on MindCube-tiny and from 33.7\% to 26.8\% on MMSI-Bench.
Although QA alone improves over the baseline in several cases, \ourmethod outperforms both variants in most settings.
This indicates that the QA step supports both view seeking and subsequent final reasoning.

\noindent\textbf{Role of the Action Interface.}
Table~\ref{tab:s2_viewpoint_choice} compares alternative strategies for determining the viewpoint.
The 6-DoF variant directly predicts a numerical camera pose, the random variant samples from the predefined action set, and view interpolation synthesizes an intermediate observation between neighboring views.
Across all four models, 6-DoF prediction underperforms our action-selection interface, while random actions and view interpolation provide limited gains overall.
Notably, view interpolation only surpasses our method for Qwen3-VL-8B.
These results support structured action selection as a more reliable interface between semantic planning and executable camera motion, and suggest that the gains of \ourapproach come from seeking question-relevant viewpoints rather than simply adding denser or more continuous observations.

\begin{table}[t]
\centering
\begin{minipage}[t]{.49\linewidth}
\vspace{0pt}
    \centering
    \small
    \setlength{\tabcolsep}{11pt}
    \caption{\textbf{Question analysis (QA) step ablation.} 
    QA-skip plans views without the analysis; QA-only augments final VQA with analysis alone.
    }
    \label{tab:per_module_ablation}
    \resizebox{\linewidth}{!}{
    \begin{tabular}{l|l|cc}
        \toprule
        \textbf{Model} & \textbf{Setting} & \textbf{MindCube-tiny} & \textbf{MMSI-Bench} \\
        \midrule

        \multirow{4}{*}{Qwen3-VL-4B}
        & Baseline                     & 26.0 & 27.9 \\
        & \ourmethod                   & \textbf{38.9} & \textbf{33.7} \\
        & QA-skip       & 30.0 & 26.8 \\
        & QA-only       & 34.8 & 33.3 \\
        \midrule

        \multirow{4}{*}{Qwen3-VL-8B}
        & Baseline                     & 31.3 & 30.1 \\
        & \ourmethod                   & \textbf{31.6} & 30.1 \\
        & QA-skip       & 31.2 & \textbf{31.6} \\
        & QA-only       & 27.5 & 28.5 \\
        \midrule

        \multirow{4}{*}{Qwen3.6-27B}
        & Baseline                     & 63.0 & 42.3 \\
        & \ourmethod                   & \textbf{68.8} & \textbf{44.8} \\
        & QA-skip       & 65.0 & 43.3 \\
        & QA-only       & 63.4 & 42.2 \\
        \midrule

        \multirow{4}{*}{Gemma-4-31B}
        & Baseline                     & 53.5 & 36.8 \\
        & \ourmethod                   & \textbf{61.9} & 37.1 \\
        & QA-skip       & 58.7 & \textbf{37.3} \\
        & QA-only       & 57.9 & 36.5 \\
        \bottomrule
    \end{tabular}
    }
\end{minipage}\hfill
\begin{minipage}[t]{.49\linewidth}
\vspace{0pt}
    \centering
    \small
    \setlength{\tabcolsep}{14pt}
    \caption{\textbf{Action interface ablation.} We compare action selection, 6-DoF prediction, random actions and view interpolation on MindCube-tiny.}
    \label{tab:s2_viewpoint_choice}
    \resizebox{0.994\linewidth}{!}{
    \begin{tabular}{l|l|cc}
    \toprule
    \textbf{Model} & \textbf{Setting} & \textbf{Acc.} & \textbf{$\Delta$} \\
    \midrule
    \multirow{4}{*}{Qwen3-VL-4B} & Action selection (ours) & \textbf{38.9} & -- \\
    & 6-DoF prediction & 34.6 & $-4.3$ \\
    & Random action & 33.0 & $-5.9$ \\
    & View interpolation & 32.1 & $-6.8$ \\
    \midrule
    \multirow{4}{*}{Qwen3-VL-8B} & Action selection (ours) & 31.6 & -- \\
    & 6-DoF prediction & 31.2 & $-0.4$ \\
    & Random action & 31.8 & $+0.2$ \\
    & View interpolation & \textbf{32.4} & $+0.8$ \\
    \midrule
    \multirow{4}{*}{Qwen3.6-27B} & Action selection (ours) & \textbf{68.8} & -- \\
    & 6-DoF prediction & 67.4 & $-1.4$ \\
    & Random action & 63.1 & $-5.7$ \\
    & View interpolation & 64.3 & $-4.5$ \\
    \midrule
    \multirow{4}{*}{Gemma-4-31B} & Action selection (ours) & \textbf{61.9} & -- \\
    & 6-DoF prediction & 61.5 & $-0.4$ \\
    & Random action & 55.2 & $-6.7$ \\
    & View interpolation & 54.6 & $-7.3$ \\
    \bottomrule
    \end{tabular}
    }
\end{minipage}
\end{table}

\begin{table}[t]
    \centering
    \setlength{\tabcolsep}{8pt}
    \caption{\textbf{Final VQA context ablation.} 
    Checkmarks indicate the component provided to the original VQA.
    Bold and italic denote the best and second-best average accuracy within each model, respectively.
    The final two columns report the absolute and relative improvements over its baseline.
    }
    \label{tab:ivs_components}
    \resizebox{\linewidth}{!}{
        \begin{tabular}{lccc|ccc|ccc}
        \toprule
        \multirow{2.5}{*}{\textbf{Model}} & \multicolumn{3}{c}{\textbf{Augmentation}} & \multicolumn{3}{c}{\textbf{Benchmark}} & \multirow{2.5}{*}{\textbf{Avg.}} & \multirow{2.5}{*}{\textbf{Abs $\Delta$}} & \multirow{2.5}{*}{\textbf{Relative $\Delta$}}\\
        \cmidrule(lr){2-4}\cmidrule(lr){5-7}
        & \shortstack{View} & \shortstack{Analysis} & \shortstack{Guidance} & MindCube-tiny & MMSI-Bench & OmniSpatial & \\
        \midrule
        \multirow{6}{*}{Qwen3-VL-4B}
        &  &  &  & 26.0 & 27.9 & 42.1 & 32.0 & -- & -- \\
        &  & \checkmark & \checkmark & 34.5 & 32.1 & 43.1 & 36.6 & +4.6 & +14.4 \\
        & \checkmark &  &  & 29.7 & 32.7 & 43.5 & 35.3 & +3.3 & +10.3 \\
        & \checkmark & \checkmark &  & 38.2 & 34.0 & 42.4 & \textit{38.2} & +6.2 & +19.4 \\
        & \checkmark &  & \checkmark & 36.7 & 31.4 & 42.1 & 36.7 & +4.7 & +14.7 \\
        & \checkmark & \checkmark & \checkmark & 38.9 & 33.7 & 47.2 & \textbf{39.9} & \textbf{\textcolor{darkblue}{+7.9}} & \textbf{\textcolor{darkgreen}{+24.7}}\\
        \midrule
        \multirow{6}{*}{Gemma-4-31B}
        &  &  &  & 53.5 & 36.8 & 51.7 & 47.3 & -- & -- \\
        &  & \checkmark & \checkmark & 56.9 & 37.6 & 53.6 & 49.4 & +2.1 & +4.4 \\
        & \checkmark &  &  & 56.8 & 34.5 & 55.4 & 48.9 & +1.6 & +3.4 \\
        & \checkmark & \checkmark &  & 59.3 & 36.4 & 55.3 & \textit{50.3} & +3.0 & +6.3 \\
        & \checkmark &  & \checkmark & 60.9 & 35.4 & 51.7 & 49.3 & +2.0 & +4.2 \\
        & \checkmark & \checkmark & \checkmark & 61.9 & 37.1 & 55.1 & \textbf{51.4} & \textbf{\textcolor{darkblue}{+4.1}} & \textbf{\textcolor{darkgreen}{+8.7}} \\
        \bottomrule
        \end{tabular}
    }
\end{table}

\noindent\textbf{Role of the Visual and Textual Context.}
Table~\ref{tab:ivs_components} examines how the synthesized view $v_s$, question analysis $c_{\mathrm{ana}}$, and reasoning guidance $c_{\mathrm{guide}}$ contribute to the final VQA step. 
All preceding steps are kept fixed, with only the context supplied to the final VQA being varied.
Using the synthesized view alone improves the average accuracy across the three benchmarks for all four models.
Combining the view with textual context generally yields further gains, while the full combination consistently outperforms the setting using only question analysis and reasoning guidance (row 2).
These results indicate that visual and textual context play complementary roles: the synthesized view provides additional spatial evidence, while the analysis and guidance preserve the observation objective and help the model interpret this evidence during the final reasoning.
\begin{figure}[t]
    \centering
    \begin{subfigure}[t]{0.31\linewidth}
        \centering
        \includegraphics[trim=8 8 8 8, clip, width=\linewidth]{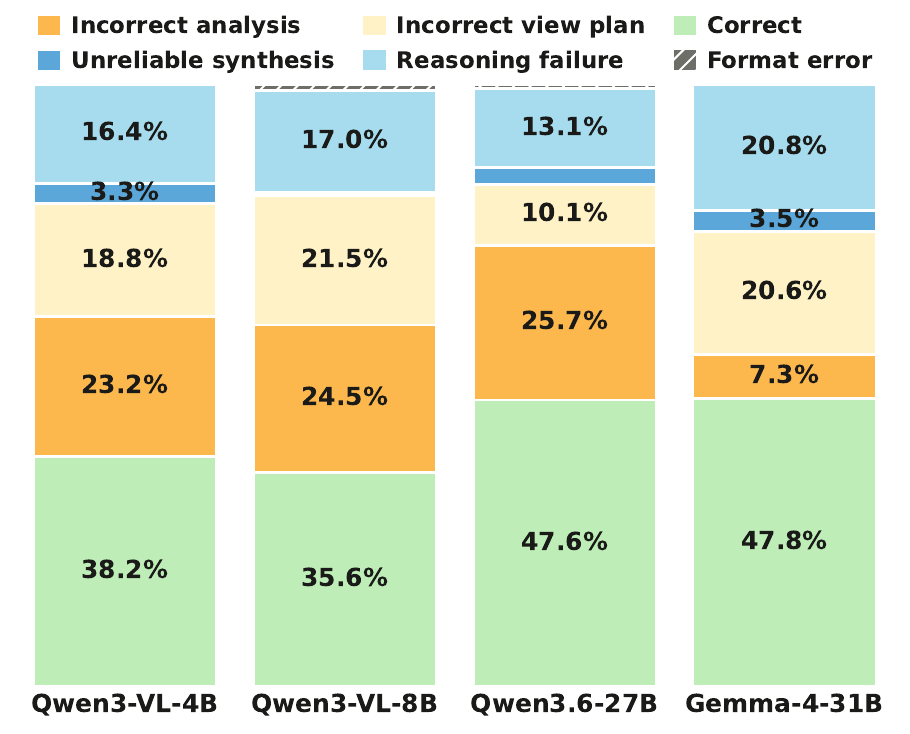}
    \end{subfigure}
    \hfill
    \begin{subfigure}[t]{0.68\linewidth}
        \centering
        \includegraphics[width=\linewidth]{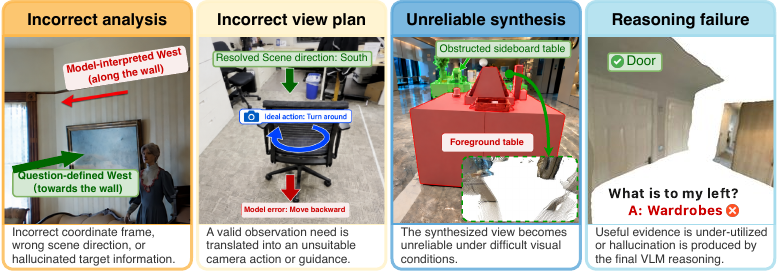}
    \end{subfigure}
    \caption{
        \textbf{Major failure types of \ourmethod.}
        \textit{Left}: Macro-averaged failure analysis.
        \textit{Right}: Representative failure cases by type.
        More detailed visualizations of these representative failure cases are provided in the Appendix.
    }
    \label{fig:failure_analysis}
\end{figure}

\subsection{Discussion}
Our training-free model-agnostic framework enables a diagnostic analysis of where current VLMs + \ourapproach still fail.
We audit 600 incorrect predictions across four models and three multi-view benchmarks. 
Figure~\ref{fig:failure_analysis} summarizes the estimated prevalence of different failure types and representative cases. 
Detailed categorization and estimation procedures are provided in the Appendix.

\noindent\textcolor{orange}{\textbf{What Limits View Seeking?}}
The dominant failures arise in the viewpoint-grounded reasoning stage.
\emph{Incorrect analysis} and \emph{incorrect view plan} together account for most residual errors, indicating that deciding \emph{what should be seen} and \emph{how to reach it} remains the main bottleneck.
As illustrated in Figure~\ref{fig:failure_analysis} (right), these failures reflect errors in analyzing the spatial context or in translating a valid observation need into an executable camera action.
These results suggest that effective viewpoint seeking depends on both accurate question interpretation and reliable grounding from semantic intent to geometric action. This trend is further supported by the cross-source results in Table~\ref{tab:xfer_ablation}.

\noindent\textcolor{cyan}{\textbf{What Limits Reasoning from the Sought View?}}
Failures at this stage mainly arise from \emph{unreliable synthesis} or \emph{reasoning failure}.
Unreliable synthesis is rare and mostly occurs under challenging visual conditions, such as reflections, transparency, severe occlusion, close-up views, darkness, blur, or low resolution.
This suggests that \ourapproach tolerates imperfect rendering, as long as the synthesized view preserves sufficient geometric and semantic evidence for reasoning.
Reasoning failures occur when the final VLM under-utilizes useful evidence in the sought view or produces an inconsistent reasoning trace.
Thus, the sought view provides additional visual evidence to support cross-view alignment, while reliable spatial reasoning remains necessary for accurate answering.
\section{Conclusion}
We formulated \ourapproach, a new reasoning paradigm that seeks question-relevant viewpoints to make spatial evidence distributed across views more directly observable.
This paradigm alleviates the burden on semantic textual reasoning for global scene understanding through question-relevant, evidence-seeking and reasoning-oriented visual evidence acquisition.
To instantiate \ourapproach, we proposed \ourmethod, which integrates a VLM with a 3D foundation model through a viewpoint-grounded reasoning stage and a geometry-grounded evidence augmentation stage.
Across five benchmarks and six VLMs, this training-free and model-agnostic framework consistently improves spatial reasoning.
The results suggest that more robust multi-view spatial understanding does not always require strengthening reasoning over fixed observations; rather, seeking question-relevant visual evidence supports cross-view alignment and enhances subsequent reasoning.

\noindent\textbf{Limitation \& Future work.} 
As a training-free framework, \ourmethod may yield rollouts of varying quality depending on the capabilities of the underlying VLM and 3DFM.
Future work could leverage high-quality rollouts for end-to-end optimization toward more reliable spatial reasoning.

\subsection*{Acknowledgment}
This research was undertaken with the assistance of resources from Monash University and the National Computational Infrastructure (NCI Australia) allocation scheme. NCI is an NCRIS-enabled capability supported by the Australian Government.

\bibliography{refs}
\bibliographystyle{unsrtnat}

\appendix
\section*{Appendix}

The appendix is organized as follows:
\begin{itemize}
    \item Section~\ref{sec:method_details} provides additional design and implementation details of \ourmethod, including the action set, view synthesis pipeline, gravity alignment, reconstruction backend, and prompt templates.
    \item Section~\ref{sec:benchmark_details} provides detailed descriptions of the evaluation benchmarks and subset selection.
    \item Section~\ref{sec:qualitative_analysis} presents additional qualitative results and the detailed failure analysis protocol.
\end{itemize}

\section{More Details of \ourmethod}
\label{sec:method_details}
This section provides additional details of \ourmethod, covering the action space, view synthesis procedure, gravity alignment study, and prompt templates.

\subsection{Action Set for View Planning}
We provide the complete action set used for view planning in Table~\ref{tab:action_set}, together with the corresponding magnitude levels.
We adopt the camera convention $+\mathrm{yaw}$ for leftward rotation, $+\mathrm{pitch}$ for upward tilt, and $+x$, $+y$, and $+z$ for rightward, downward, and forward translation, respectively.
Rotation actions modify only the viewing direction, whereas translation actions move the camera by a distance specified in scene units.
For hybrid transformations, orbit actions rotate the camera around the centroid of the reconstructed point cloud, while bird's-eye and worm's-eye views introduce additional pitch rotations to obtain elevated or lowered viewpoints.
Since the reconstructed point cloud is not metrically scaled, absolute translation distances are not directly transferable across scenes.
We therefore define one scene unit $s$ as the mean Euclidean distance from the reference camera center to the remaining input camera centers, providing a scene-adaptive scale that reflects the spatial spread of the observed viewpoints.

\begin{table}[ht]
\centering
\small
\setlength{\tabcolsep}{12pt}
\caption{\textbf{Action set for view planning.}
Each semantic action is deterministically mapped to a geometric transformation with three magnitude levels (small, medium, and large), except for \textit{turn around}, which uses a fixed $180^\circ$ rotation. Translation magnitudes are expressed in scene unit $s$.}
\label{tab:action_set}
\resizebox{0.88\linewidth}{!}{
\begin{tabular}{lllccc}
    \toprule
    Category & Action & Transformation & Small & Medium & Large \\
    \midrule
    \multirow{5}{*}{\textbf{Rotation}}
    & Pan left
    & $+\Delta\mathrm{yaw}$
    & $30^\circ$ & $60^\circ$ & $90^\circ$ \\
    & Pan right
    & $-\Delta\mathrm{yaw}$
    & $30^\circ$ & $60^\circ$ & $90^\circ$ \\
    & Tilt up
    & $+\Delta\mathrm{pitch}$
    & $30^\circ$ & $60^\circ$ & $90^\circ$ \\
    & Tilt down
    & $-\Delta\mathrm{pitch}$
    & $30^\circ$ & $60^\circ$ & $90^\circ$ \\
    & Turn around
    & $\Delta\mathrm{yaw}$
    & \multicolumn{3}{c}{$180^\circ$} \\
    \midrule
    \multirow{6}{*}{\textbf{Translation}}
    & Move right
    & $+\Delta x$
    & $0.40s$ & $0.80s$ & $1.40s$ \\
    & Move left
    & $-\Delta x$
    & $0.40s$ & $0.80s$ & $1.40s$ \\
    & Pedestal down
    & $+\Delta y$
    & $0.30s$ & $0.55s$ & $0.90s$ \\
    & Pedestal up
    & $-\Delta y$
    & $0.30s$ & $0.55s$ & $0.90s$ \\
    & Move forward
    & $+\Delta z$
    & $0.40s$ & $0.80s$ & $1.40s$ \\
    & Move backward
    & $-\Delta z$
    & $0.40s$ & $0.80s$ & $1.40s$ \\
    \midrule
    \multirow{6}{*}{\textbf{Hybrid}}
    & Orbit left
    & Orbit around centroid, left
    & $30^\circ$ & $60^\circ$ & $90^\circ$ \\
    & Orbit right
    & Orbit around centroid, right
    & $30^\circ$ & $60^\circ$ & $90^\circ$ \\
    & Orbit up
    & Orbit around centroid, upward
    & $30^\circ$ & $60^\circ$ & $90^\circ$ \\
    & Orbit down
    & Orbit around centroid, downward
    & $30^\circ$ & $60^\circ$ & $90^\circ$ \\
    & Bird's-eye view
    & $+\Delta\mathrm{pitch}$
    & $40^\circ$ & $55^\circ$ & $70^\circ$ \\
    & Worm's-eye view
    & $-\Delta\mathrm{pitch}$
    & $40^\circ$ & $55^\circ$ & $70^\circ$ \\
    \bottomrule
\end{tabular}
}
\end{table}

\subsection{View Synthesis Pipeline}
Given the target camera extrinsics determined above, we further specify the camera intrinsics for view synthesis.
In particular, the field of view (FoV) is chosen to preserve sufficient scene coverage after the planned camera transformation.
For rotation and translation actions, we use a fixed FoV of $120^\circ$ to provide a sufficiently wide observation.
For hybrid transformations, which can substantially change the camera position relative to the scene, we adapt the FoV to maintain coverage of the reconstructed scene.
For example, a bird's-eye transformation may change the apparent scale of a compact scene, and the FoV is adjusted accordingly to retain informative coverage.
The output follows the aspect ratio of the reference view, while reprojection is performed at twice the target width and height before downsampling to the resolution of the reference view.

Table~\ref{tab:render_fov} illustrate the effect of the FoV while keeping the upstream viewpoint-grounded reasoning fixed. Reducing the diagonal FoV from $120^\circ$ to $80^\circ$ lowers accuracy by 0.7 and 2.5 \% for Qwen3.6-27B and Gemma-4-31B, respectively. This suggests that FoV affects how much spatial context can be captured within the synthesized view, which is particularly important for complex real-world scenes where relevant evidence may span a broader spatial region.

\begin{figure}[t]
    \centering
    \includegraphics[width=\linewidth]{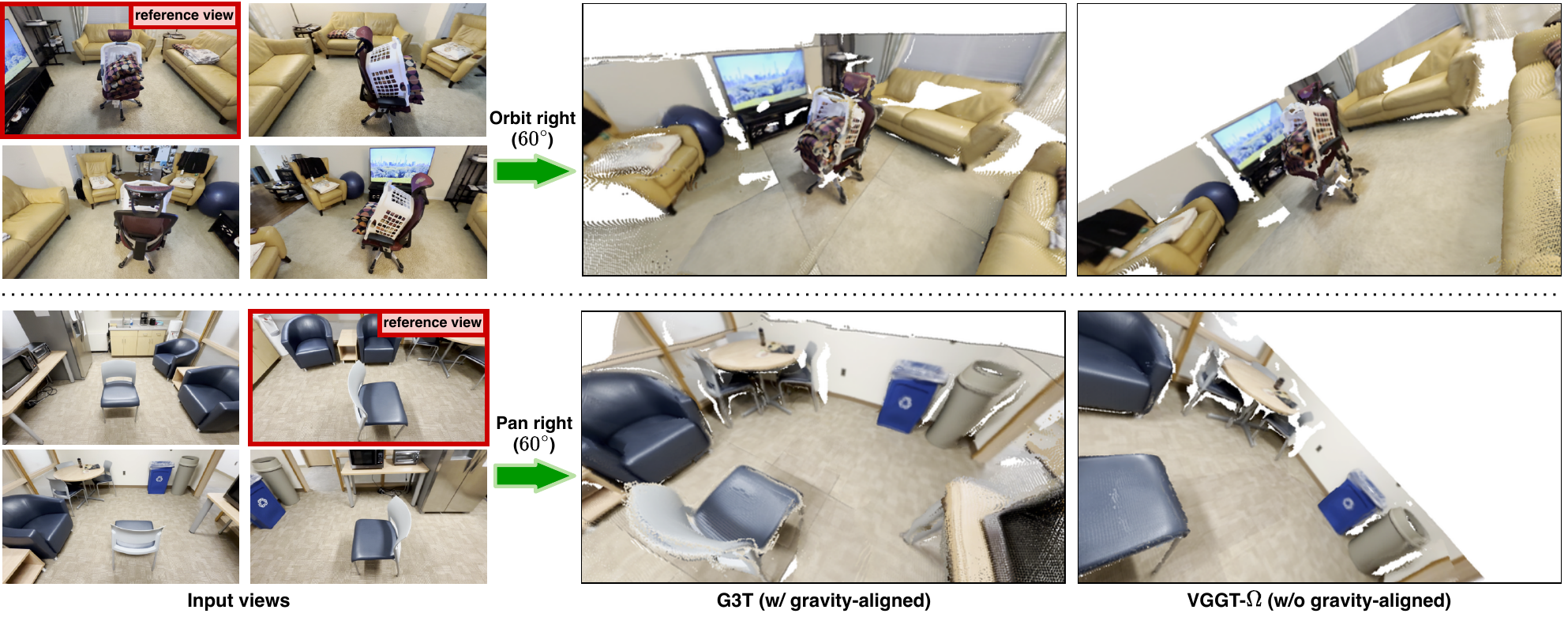}
    \caption{\textbf{Effect of gravity alignment on view synthesis.}
    Given the same input views, reference view, and planned camera action, we compare view synthesis using a gravity-aligned G3T reconstruction and a non-gravity-aligned VGGT-$\Omega$ reconstruction. Gravity alignment provides a ground-consistent coordinate frame for executing relative camera actions, while an unaligned reconstruction can cause the realized viewpoint to deviate from the intended ground-relative motion.}
    \label{fig:gravity_compare}
\end{figure}

Our action interface assumes a ground-aligned coordinate frame, as camera motions such as forward, backward, and lateral translation are defined relative to the ground plane.
Accordingly, the reconstructed scene should be aligned with gravity before applying the planned camera transformation.
Figure~\ref{fig:gravity_compare} illustrates this effect by applying the same planned action under different reconstruction configurations, showing how gravity alignment influences the realized camera motion.

\begin{table}[t]
    \centering
    \caption{Field-of-view ablation on MindCube.}
    \label{tab:render_fov}
    \resizebox{0.5\linewidth}{!}{
    \begin{tabular}{llcc}
    \toprule
    \textbf{Model} & \textbf{FoV} & \textbf{MindCube-tiny} & $\Delta$ \\
    \midrule
    \multirow{2}{*}{Qwen3.6-27B} & $120^\circ$ (ours) & \textbf{68.8} & -- \\
    & $80^\circ$ & 68.1 & $-0.7$ \\
    \midrule
    \multirow{2}{*}{Gemma-4-31B} & $120^\circ$ (ours) & \textbf{61.9} & -- \\
    & $80^\circ$ & 59.4 & $-2.5$ \\
    \bottomrule
    \end{tabular}
    }
\end{table}

\noindent\textbf{Reconstruction Backend.}
Table~\ref{tab:recon_backend_ab} compares G3T~\citep{kani2026g3t} with VGGT-$\Omega$~\citep{wang2026vggt} combined with GeoCalib~\citep{veicht2024geocalib}, representing two routes to the gravity-aligned reconstruction required by our action interface.
G3T directly produces a gravity-aligned reconstruction, whereas the alternative first reconstructs the scene with VGGT-$\Omega$ and then uses GeoCalib on the reference view to estimate the gravity direction for alignment.
Both backends yield comparable downstream performance.
We use G3T as the default backend, as it integrates gravity alignment directly into reconstruction and avoids the additional calibration step, while the VGGT-$\Omega$ + GeoCalib pipeline requires approximately $1.4\times$ the inference time.
\begin{table}[ht]
    \centering
    \caption{Reconstruction backends ablation.}
    \label{tab:recon_backend_ab}
    \resizebox{0.7\linewidth}{!}{
    \begin{tabular}{llccc}
        \toprule
        \textbf{Model} & \textbf{Backend} & \textbf{MindCube-tiny} & \textbf{MMSI-Bench} & \textbf{Avg.} \\
        \midrule
        \multirow{2}{*}{Qwen3-VL-4B} & G3T & \textbf{38.9} & \textbf{33.7} & \textbf{36.3} \\
        & VGGT-$\Omega$ + GeoCalib & 37.4 & 30.7 & 34.1 \\
        \midrule
        \multirow{2}{*}{Qwen3-VL-8B} & G3T & 31.6 & \textbf{30.1} & \textbf{30.9} \\
        & VGGT-$\Omega$ + GeoCalib & 31.6 & 29.9 & 30.8 \\
        \midrule
        \multirow{2}{*}{Qwen3.6-27B} & G3T & \textbf{68.8} & \textbf{44.8} & \textbf{56.8} \\
        & VGGT-$\Omega$ + GeoCalib & 67.2 & 42.2 & 54.7 \\
        \midrule
        \multirow{2}{*}{Gemma-4-31B} & G3T & 61.9 & \textbf{37.1} & 49.5 \\
        & VGGT-$\Omega$ + GeoCalib & \textbf{63.5} & 35.6 & \textbf{49.6} \\
        \bottomrule
    \end{tabular}
    }
\end{table}

\subsection{Prompt Templates}

Prompts~\ref{prompt:s1}--\ref{prompt:s4fb} provide the prompt templates used throughout \ourmethod pipeline.

Prompt~\ref{prompt:s1} is used for viewpoint analysis.
Given the input views $\mathcal{V}$ and question $q$, the VLM produces a structured, viewpoint-centric analysis $c_{\mathrm{ana}}$ that identifies the reference view, anchor entity, frame of reference, and observation need.
At this step, the model does not answer the question or select a camera action.

Prompt~\ref{prompt:s2} is used for view planning.
Conditioned on the input views, question, and viewpoint analysis, the VLM selects a relative camera action $\Delta v$ with respect to the reference view and generates the reasoning guidance $c_{\mathrm{guide}}$ for subsequent answering.
The action is represented by a predefined action type and magnitude level, rather than direct numerical camera parameters.

Prompt~\ref{prompt:s4} is used for final VQA.
The synthesized view $v_s$ is appended to the original input views, and the VLM receives the original question together with $c_{\mathrm{ana}}$ and $c_{\mathrm{guide}}$ as additional context for final answering.

Prompt~\ref{prompt:s4fb} defines the fallback setting triggered by a format error.
In this case, the VLM answers using only the original input views together with the available textual context, namely the question analysis and, if available, the reasoning guidance.

\section{Benchmark Details}
\label{sec:benchmark_details}
This section provides additional descriptions of the benchmarks summarized in Table~\ref{tab:main} of the main paper, including the evaluated subsets and our subset selection criteria.

\subsection{Benchmark Descriptions}
\paragraph{MindCube-tiny.}
MindCube evaluates spatial mental modeling from limited multi-view observations, covering spatial relations that require reasoning about object positions, orientations, and hypothetical viewpoint changes~\citep{wang2026mindcube}.
We evaluate on the MindCube-tiny split and report its Rotation, Among, and Around subsets.

\paragraph{MMSI-Bench.}
MMSI-Bench evaluates multi-view spatial intelligence over real-world scenes, covering positional relations among cameras, objects, and regions, as well as attribute, motion, and multi-step reasoning~\citep{yang2026mmsi}.
We focus on the subsets centered on static multi-view spatial reasoning and exclude the Motion and Multi-Step Reasoning subsets. The former involves temporal changes across observations, while the latter places greater emphasis on compositional reasoning over multiple spatial steps.
Both introduce additional reasoning factors beyond the setting targeted by \ourapproach, which focuses on seeking question-relevant visual evidence when spatial information is fragmented across sparse observations or difficult to express and resolve through language alone.

\paragraph{BLINK.}
BLINK is a diagnostic benchmark covering wide perception-demanding visual tasks, including multi-view reasoning, relative depth, visual correspondence, and spatial relations~\citep{fu2024blink}.
We use only its Multi-view Reasoning subset, which requires reasoning across observations from different viewpoints and is therefore most aligned with our multi-view spatial setting.

\paragraph{OmniSpatial.}
OmniSpatial evaluates spatial reasoning across four broad dimensions: dynamic reasoning, complex spatial logic, spatial interaction, and perspective taking~\citep{jia2026omnispatial}.
We include only the Perspective Taking subset, which evaluates egocentric, allocentric, and hypothetical viewpoint reasoning from a single observed scene.
The remaining categories primarily emphasize temporal dynamics, interaction and planning, or abstract geometric and pattern reasoning, and are not designed around multi-view observations of real-world scenes.

\paragraph{SPINBench.}
SPINBench is a diagnostic benchmark for spatial reasoning under perspective and geometric transformations, covering translation, rotation, relative pose, and viewpoint change~\citep{zhang2026spinbench}.
We additionally evaluate on the Dynamic Rotation and Dynamic Translation subsets, where an object or subject undergoes controlled rotation or translation while the surrounding scene largely remains static.
These subsets provide a limited extension beyond our core static multi-view setting by testing whether \ourapproach can also support spatial reasoning when the queried entity itself changes pose or position.

\section{Additional Qualitative Analysis}
\label{sec:qualitative_analysis}
\begin{figure}[ht]
    \centering
    \includegraphics[width=\linewidth]{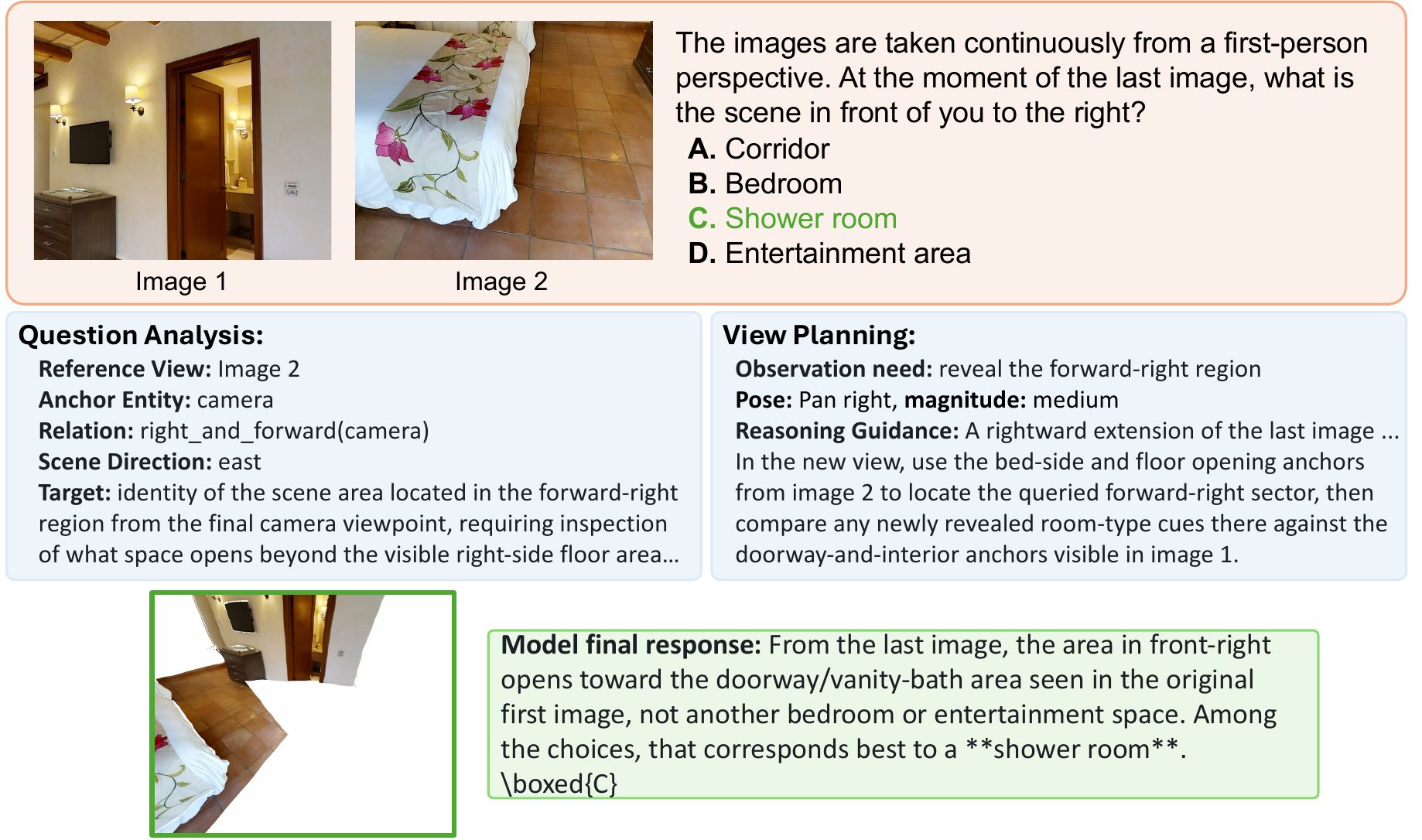}
    \caption{Qualitative example of \ourmethod with GPT-5.4.
    For clarity, we simplify the original model outputs and present only the key intermediate information relevant to the reasoning process.}
    \label{fig:case_study_1}
\end{figure}

\begin{figure}[p]
    \centering
    \includegraphics[width=\linewidth]{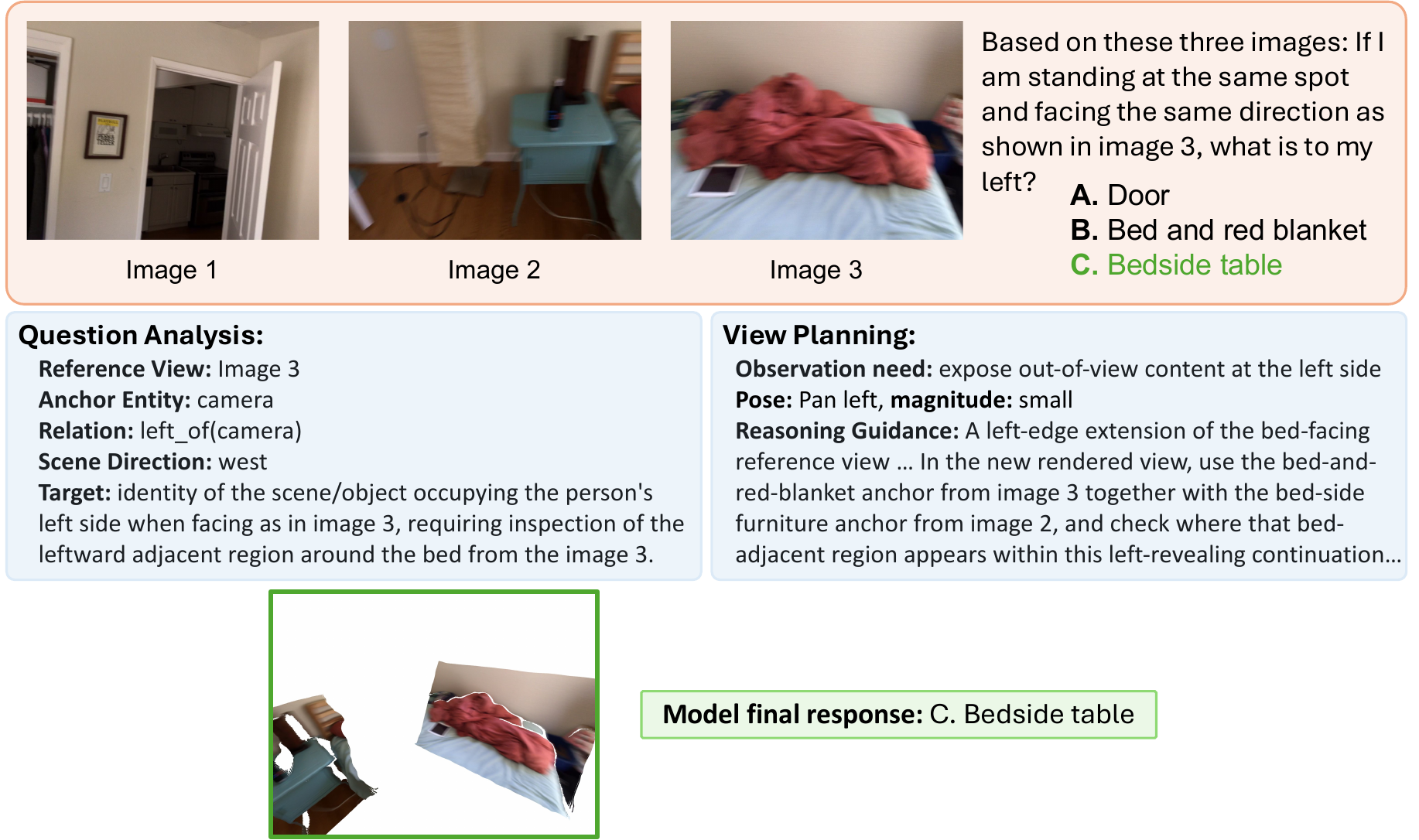}
    \caption{Qualitative example of \ourmethod with GPT-5.4.}
    \label{fig:case_study_2}
\end{figure}

\begin{figure}[p]
    \centering
    \includegraphics[width=\linewidth]{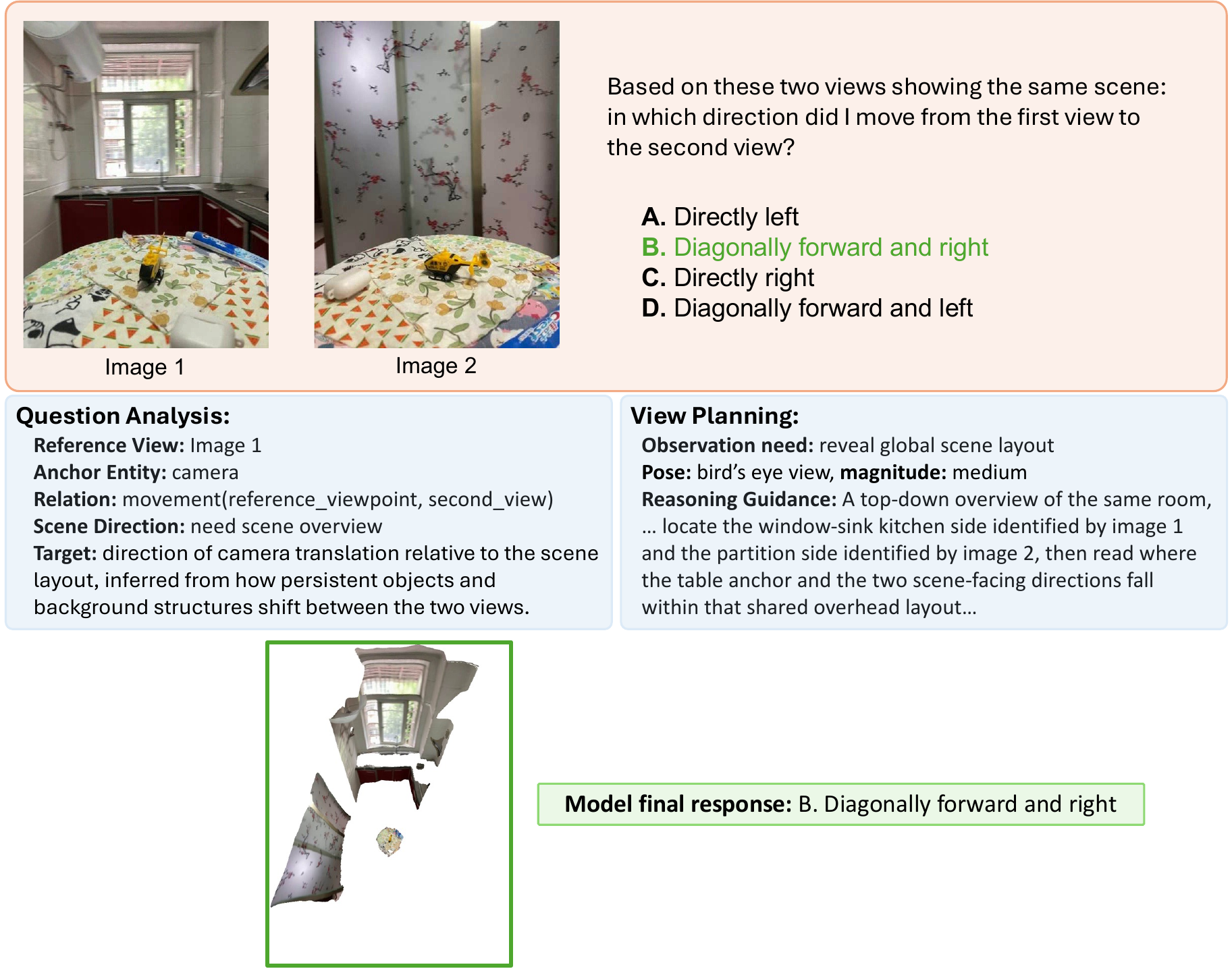}
    \caption{Qualitative example of \ourmethod with GPT-5.4.}
    \label{fig:case_study_3}
\end{figure}

\begin{figure}[p]
    \centering
    \includegraphics[width=\linewidth]{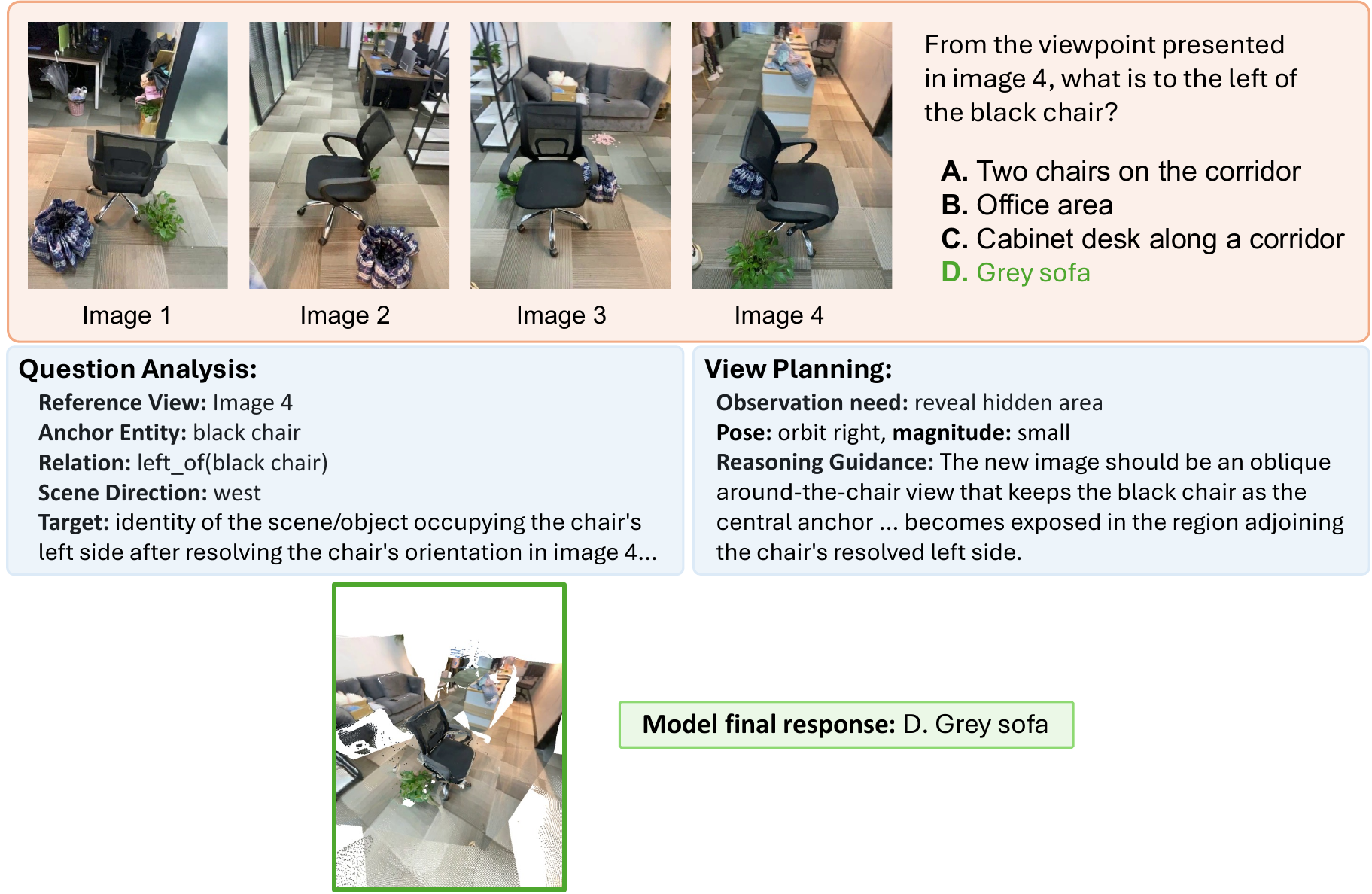}
    \caption{Qualitative example of \ourmethod with GPT-5.4.}
    \label{fig:case_study_4}
\end{figure}

This section provides additional qualitative analyses of \ourmethod, including representative success and failure cases shown in Figures~\ref{fig:case_study_1}--\ref{fig:case_study_4} .
We also provide further details of the failure analysis protocol used in the main paper.

\begin{figure}
    \centering
    \includegraphics[width=\linewidth]{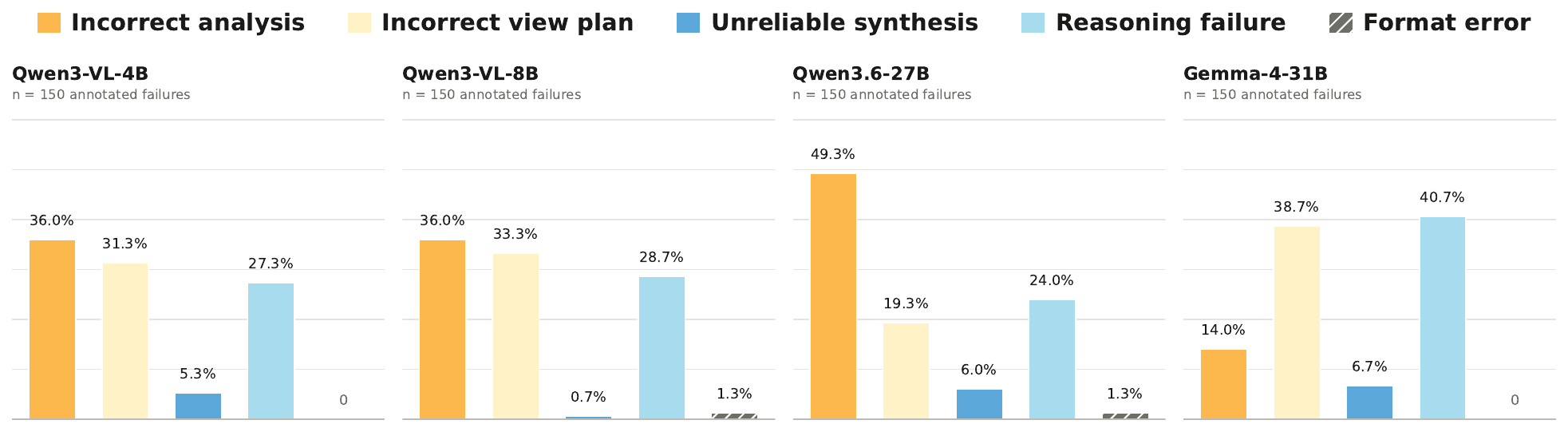}
    \caption{\textbf{Failure distribution by model.}
    For each model, we annotate 150 incorrect predictions sampled across the three multi-view benchmarks and categorize them into five failure types.
    Bars indicate the proportion of annotated failures falling into each category.
    }
    \label{fig:failure_by_model}
\end{figure}

\paragraph{Failure Analysis Protocol.}
We randomly sample 50 incorrect predictions for each of four models on each of three multi-view benchmarks, yielding 600 annotated cases in total.
Each case is assigned to one of five failure categories: \emph{Incorrect analysis}, \emph{Incorrect view plan}, \emph{Unreliable synthesis}, \emph{Reasoning failure}, or \emph{Format error}.
Figure~\ref{fig:failure_by_model} reports the resulting per-model failure distributions.
To account for differences in model accuracy, we weight the sampled proportion of each failure type by the corresponding empirical error rate and then macro-average the resulting estimates across benchmarks.
The aggregated results are reported in Figure~\ref{fig:failure_analysis} of the main paper.

\begin{figure}[p]
    \centering
    \includegraphics[width=\linewidth]{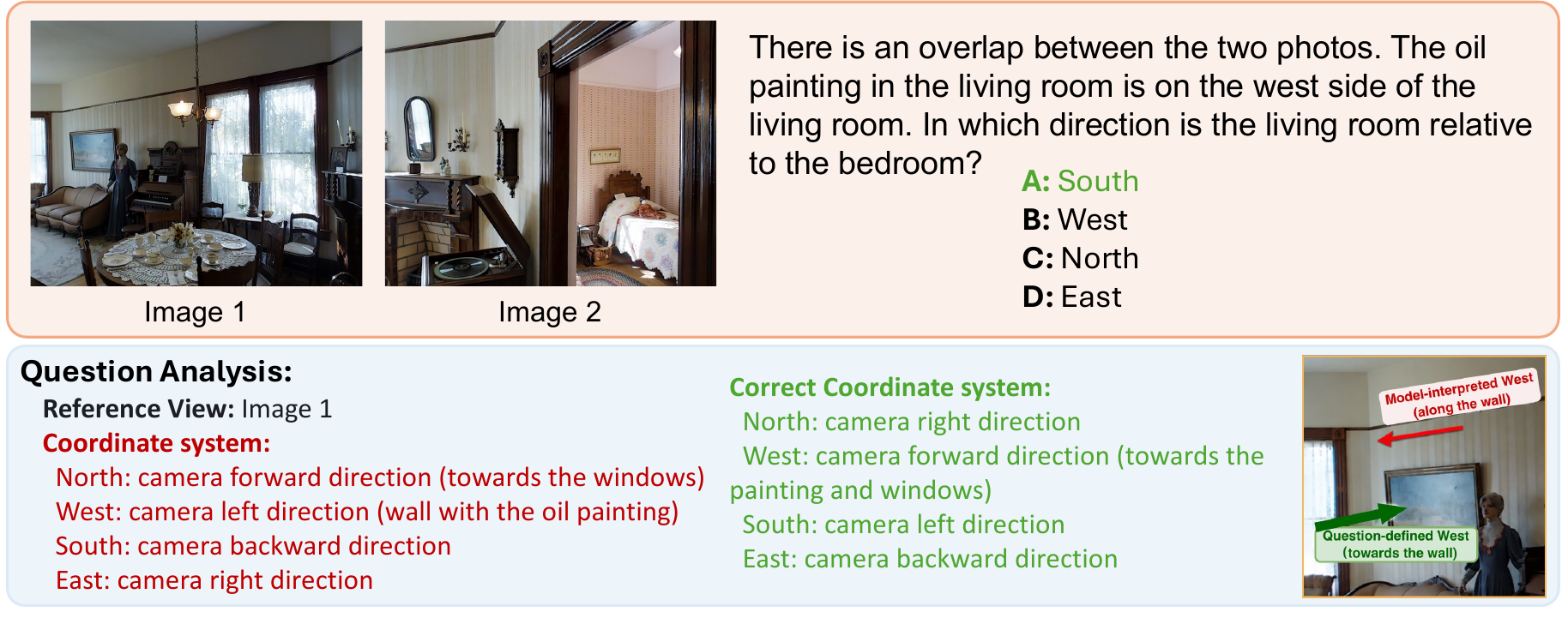}
    \caption{\textbf{Incorrect analysis.} Qualitative failure case from Qwen3.6-27B.}
    \label{fig:failure_study_1}
\end{figure}

\begin{figure}[p]
    \centering
    \includegraphics[width=\linewidth]{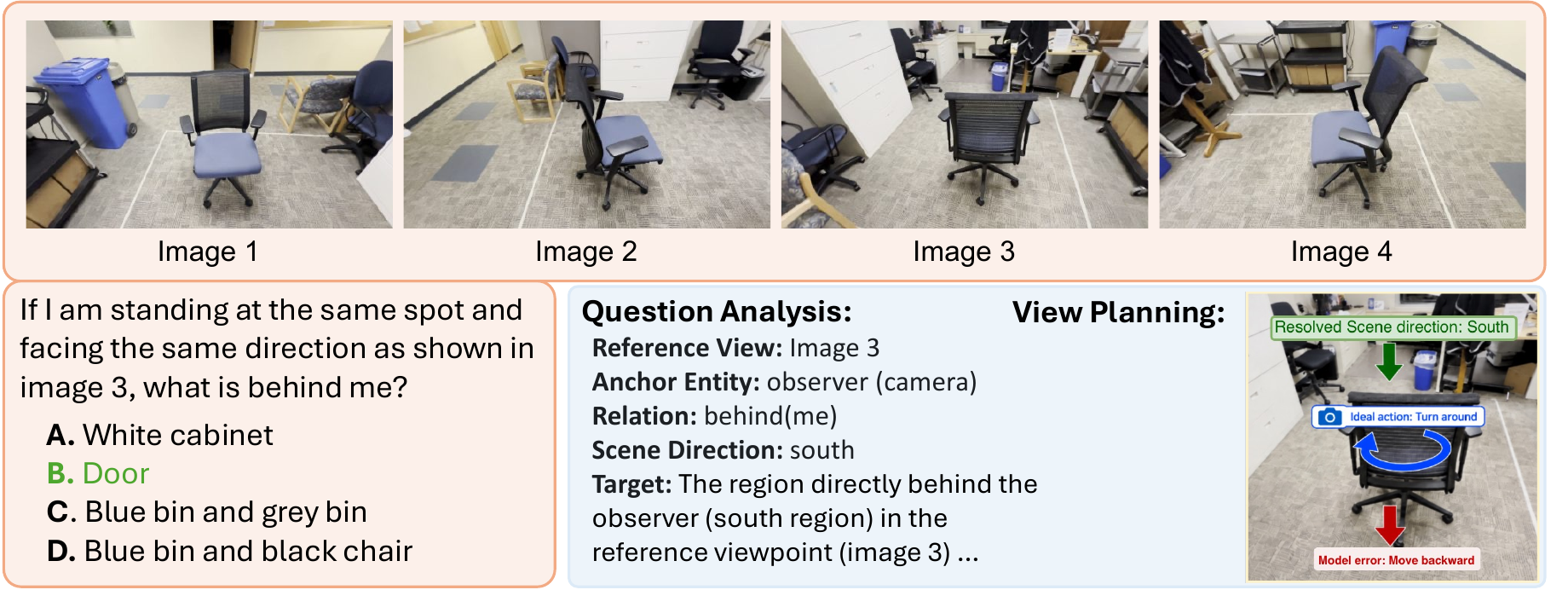}
    \caption{\textbf{Incorrect view plan.} Qualitative failure case from Qwen3-VL-4B.}
    \label{fig:failure_study_2}
\end{figure}

\begin{figure}[p]
    \centering
    \includegraphics[width=\linewidth]{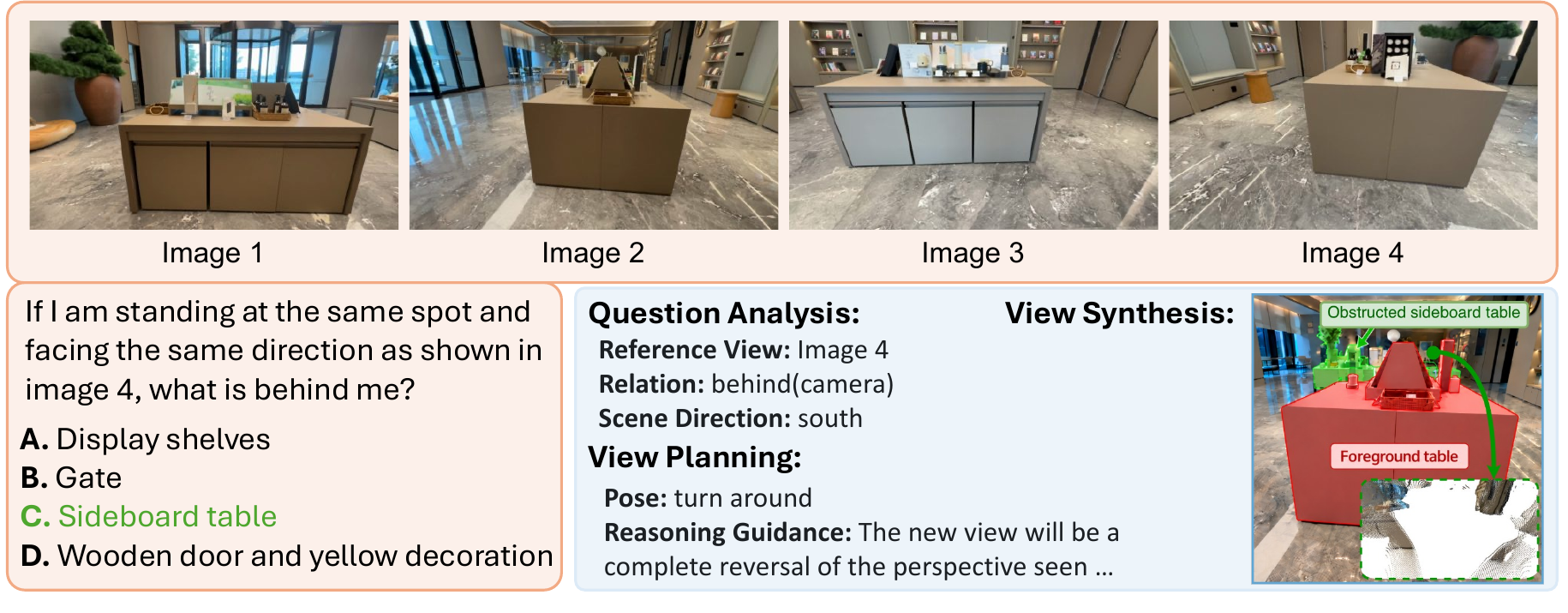}
    \caption{\textbf{Unreliable synthesis.} Qualitative failure case from Gemma-4-31B.}
    \label{fig:failure_study_3}
\end{figure}

\begin{figure}[p]
    \centering
    \includegraphics[width=\linewidth]{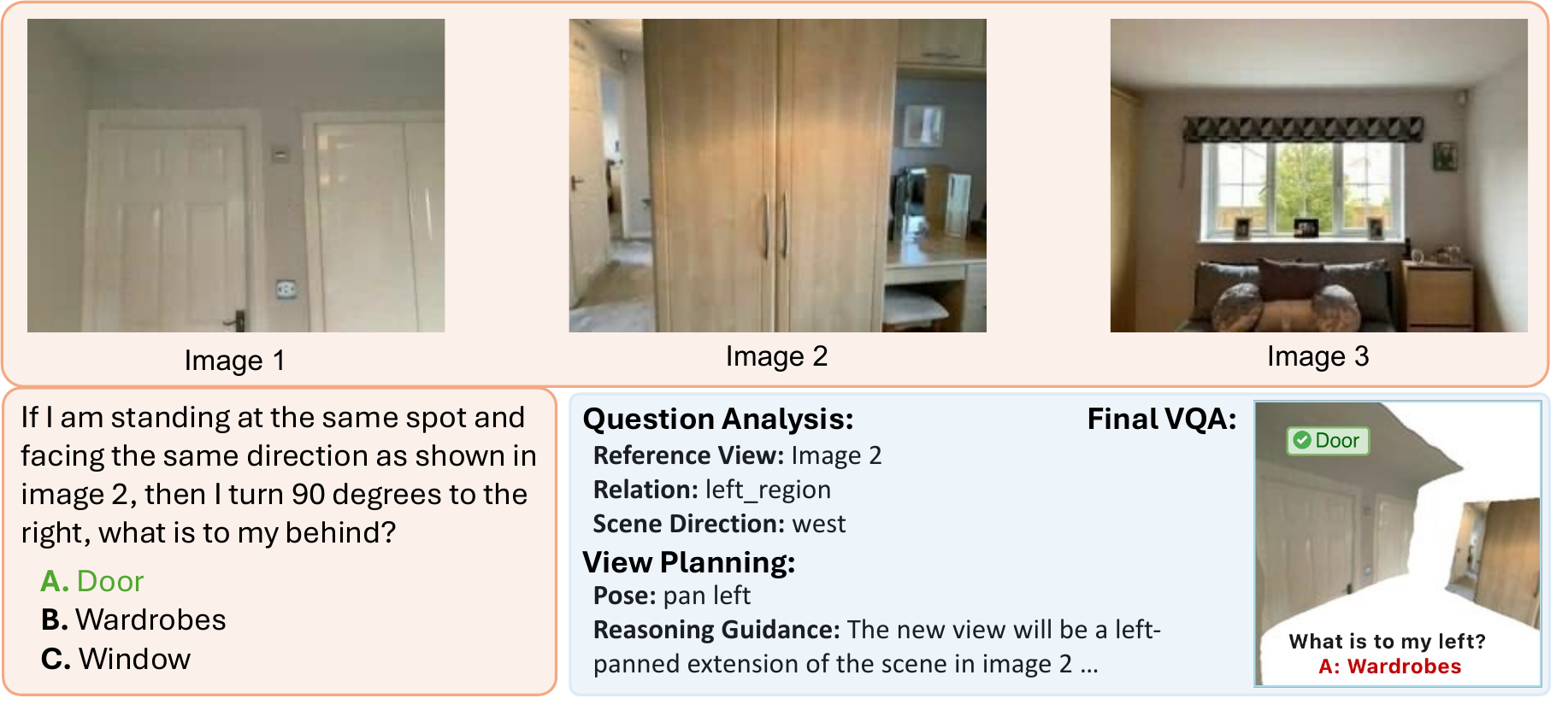}
    \caption{\textbf{Reasoning failure.} Qualitative failure case from Qwen3-VL-4B.}
    \label{fig:failure_study_4}
\end{figure}

\paragraph{Failure Case Visualizations.}
Figures~\ref{fig:failure_study_1}--\ref{fig:failure_study_4} provide detailed visualizations of the representative cases shown in Figure~\ref{fig:failure_analysis} (right), covering the four major failure types and illustrating how errors arise at different stages of the \ourmethod pipeline.

\clearpage

\clearpage

\begin{promptbox}{Stage 1 - Question Analysis}{prompt:s1}
You are an expert visual-spatial reasoning analyst.

**[CORE MISSION]**

Your sole mission is to analyze the question and construct a viewpoint-centric representation of it: the reference viewpoint to reason from, and the missing evidence and inspection region required to answer the question.

You are NOT allowed to answer the question. Your job is ONLY to:
1. Construct the local coordinate systems implied by the question.
2. Resolve any object-centric spatial relations into viewpoint-centric regions.
3. Identify what information is required and currently uncertain or missing.

**[KEY PRINCIPLE]**

The reference viewpoint is the only executable coordinate system.

Object-centric coordinates, scene-centric coordinates, and other inferred coordinate systems are intermediate reasoning tools only. All final outputs MUST be projected back into the reference viewpoint space.

**[COORDINATE SYSTEM RULES]**

By default, the reference viewpoint defines the coordinate system:
north = camera forward, west = camera left, south = camera backward, east = camera right.

When helpful (e.g., the view is oblique or the queried direction is easy to confuse), append visible scene content in parentheses to reduce 2D directional ambiguity, e.g., "north = camera forward direction (dining table in background)" or "west = camera left direction (kitchen area)". Annotate only content visible in the reference viewpoint.

If the question explicitly defines another coordinate system (e.g., "The oven is north of the sink."), infer the question-defined coordinate system before resolving spatial relations.

**[OUTPUT FORMAT]**

Return a single JSON object wrapped in ```json ... ```.

```json
{
    "question_decomposition": "...",
    "reference_view": "...",
    "reference_view_coordinate_system": {
        "north": "...",
        "west": "...",
        "south": "...",
        "east": "..."
        },
    "anchor_entity": "...",
    "entity_forward_direction": "...",
    "query_relation": "...",
    "resolved_scene_direction": "...",
    "viewpoint_inspection_target": "...",
    "target_information": "...",
    "gap_depth": "near | mid | far | unknown"
}
```

**[FIELD DEFINITIONS]**
- question_decomposition: A concise decomposition of the question.
- reference_view: The input image whose viewpoint the question explicitly refers to. Return its 1-indexed image index (1, 2, ...).
- reference_view_coordinate_system: The coordinate system based on the reference viewpoint.
    Examples: "north = camera forward direction (dining table in background), west = camera left direction, south = camera backward direction, east = camera right direction".
- anchor_entity: The entity (object or camera) whose coordinate system defines the queried spatial relation. If no object-centric frame exists, use camera.
- entity_forward_direction: Direction of the anchor entity's forward axis expressed in the reference viewpoint coordinate system. For entities without meaningful orientation (e.g. bottle, cup, ball), use camera's forward direction.
    Examples: north, east, south, west.
- query_relation: The spatial relation or target viewpoint from the question.
    Examples: left_of(<anchor_entity>), right_of(<anchor_entity>), behind(<anchor_entity>), right_and_forward(camera), viewpoint_movement(reference_view, <viewpoint_2>), object_distance_from(<anchor_entity>).
- resolved_scene_direction: For questions that require understanding the scene region of interest, choose the direction referenced by the question after resolving any anchor-centric coordinate system: north, east, south, west.
    For questions that require understanding the direction of the viewpoint movement: need_scene_overview.
- viewpoint_inspection_target: The most important scene evidence that should be inspected from the reference viewpoint to answer the question.
    Examples: left_region, right_region, region_behind_object, object_occlusion_resolution, upper_region, lower_region, horizontal_inspection, surrounding_neighbor_overview, scene_overview, scene_topdown_overview. 
    Reminder: do not use coordinate system (e.g., "north_region(background)") to describe the target region in viewpoint space.
- target_information: The exact information required to answer the question from the reference viewpoint.
    Examples:
    - identity of object occupying the queried region
    - nearest object relative to chair
    - distance ordering among nearby objects
    - direction of camera translation relative to the scene layout
    - scene overview to establish the relative position of the anchor entity and the target entity
- gap_depth: Estimate how far the missing evidence lies from the anchor entity, can consider the scale of answer options and observed scene contents.
    Examples: near (0-1m), mid (1-5m), far (5-10m), or unknown (need the entire scene overview).


**[IMPORTANT RESTRICTIONS]**

DO NOT:
- answer the question
- determine relative camera positions between views
- reference other viewpoints when describing the target information
- speculate about unseen content and coordinate system from other viewpoints

Focus only on:
- coordinate system construction from the reference viewpoint
- the sequence of camera action instructions and relation resolution in the question
- projection into reference viewpoint space
- target information analysis


**[Question]**
(*\ph{question}*)

Now analyze the question and provide the JSON output.
\end{promptbox}
\newpage
\begin{promptbox}{Stage 1 - View Planning}{prompt:s2}
You are an expert viewpoint planner.

A prior analysis has decomposed the question into a viewpoint-centric representation with:
- the reference viewpoint to start thinking from
- the absent evidence and required inspection region to reveal

**[CORE MISSION]**

Your sole mission is to plan the viewpoint change to reveal the informative evidence and compose a spatial-clue narrative for the downstream reasoner to build on.

You are NOT allowed to answer the question.
You are NOT allowed to reason across input views: all spatial reasoning stays WITHIN one view at a time, in the reference viewpoint's frame (*\textemdash{}*) never infer relative camera poses or shared layout between input views.

Your job is ONLY to determine which new viewpoint would provide sufficient evidence for reasoning.

**[KEY PRINCIPLE]**

The goal is NOT to directly choose a camera action.

The goal is:
1. Verify the observation gap from the reference viewpoint.
2. Identify the missing evidence and query region.
3. Determine the observation principle required to reveal the evidence.
4. Select the pose that best realizes that principle.
5. Choose an appropriate move magnitude.
6. Compose a spatial-clue narrative that:
   (a) predicts, at a high level, the geometric form the new view will take,
   (b) grounds each input view relevant to the question with observable orientation cues and background/scene anchors,
   (c) sets up the cross-view correspondences the reasoner needs to combine, without combining them.

**[OUTPUT FORMAT]**

Return a single JSON object wrapped in ```json ... ```.

```json
{
    "reasoning": "...",
    "missing_evidence": "...",
    "observation_principle": "...",
    "reference_view": "...",
    "pose": "<one of pan_left, pan_right, tilt_up, tilt_down, turn_around, move_forward, move_backward, move_left, move_right, pedestal_up, pedestal_down, orbit_left, orbit_right, orbit_up, orbit_down, birds_eye_view, worms_eye_view>",
    "magnitude": "<one of small, medium, large>",
    "reasoning guidance": "..."
}
```

**[FIELD DEFINITIONS]**
- reasoning:  verify the stage-1 analysis with the question from the reference viewpoint, using within-view evidence only; do not relate input views to each other here.
- missing_evidence: Based on the 'target_information', identify the specific evidence that is unavailable from the reference viewpoint.
    Examples:
    - object immediately left of the chair
    - background hidden behind the cabinet
    - relative depth ordering among nearby objects
    - surrounding scene layout or need the entire scene overview
    - target object/view after the camera action instruction (e.g., turn left and move forward) from the question
- reference_view: The verified reference viewpoint from the stage-1 analysis.
- observation_principle: Determine what visual cue would reveal the missing evidence.
    Examples:
    - expose out-of-view content
    - reveal obstructed object/background/region
    - maintain anchor visibility while changing viewpoint
    - reveal global scene layout
    - follow the rotation instruction (e.g., turn left) to inspect the target object/view
- pose: Choose the pose that best realizes the observation principle.
You MUST use the expected visual consequence described below when selecting a pose.
    (*\ph{action set}*)
    Examples:
    - off-frame content: pan / tilt
    - hidden background: move / pedestal
    - object-relative relation: orbit
    - scene layout: birds_eye_view
    - camera-motion understanding: birds_eye_view or move_backward
    Use 'resolved_scene_direction' and 'viewpoint_inspection_target' to refine the final action.
- magnitude: Consider 'gap_depth' of the target region when choosing the magnitude:
    - small (30 degrees / one step)
    - medium (60 degrees / one and a half steps)
    - large (90 degrees / two steps)
    Magnitude is approximate for translation. Do NOT estimate metric distances.
- reasoning guidance: A spatial-clue narrative for the downstream reasoner. Write ONE paragraph composed of the three parts below, in order. The reasoner will see the input views, the original question, the newly rendered view, the stage-1 analysis and this reasoning guidance.
    (1) New-viewpoint form (*\textemdash{}*) a high-level, PREDICTIVE description of the viewing geometry the new view will take (e.g., "a top-down overview of the scene", "a right-orbit side-on view around the table anchor", "a right-panned extension of the scene in image N, beyond its right edge"). Describe only the viewing geometry; do NOT invent the specific content, objects, or spatial layout that will appear.
    (2) Input-view anchors (*\textemdash{}*) for EACH input view the question actually relies on, state as OBSERVABLE FACTS about that input image: its dominant scene-facing direction and one or two background/scene anchors that identify what it is looking at (e.g., "image 1 faces the wall with the whiteboard, with the desk on its right edge", "image 2 faces the kitchen counter, with a window on its left edge"). Skip input views the question does not depend on. If the question depends on only one input view, only that one is described here.
    (3) Correspondence setup (*\textemdash{}*) name WHICH cross-view relations the reasoner should combine in the newly rendered view to answer the question, phrased as clues to combine (e.g., "locate where the whiteboard anchor from image 1 and the window anchor from image 2 fall within this top-down view, and read the two camera footprints against each other"). Point to the clues; do NOT combine them or state the conclusion.


**[IMPORTANT RESTRICTIONS]**

DO NOT:
- answer the question
- reproduce, paraphrase, or hint at any answer option, in the reasoning guidance or elsewhere
- use motion verbs applied to the camera or objects when writing the reasoning guidance (e.g., "the camera moves right", "the object shifts forward", "turned left by ~45 degrees") or otherwise describe the transformation between views as an action, direction, or trajectory
- quantify rotations, translations, degrees, steps, or metric distances in the reasoning guidance
- infer relative camera positions or any spatial relations between input views
- reference other viewpoints when describing the target information
- speculate about unseen content and coordinate system from other viewpoints

Focus only on:
- understand the coordinate system from the reference viewpoint
- verify the stage-1 analysis with the question
- explain the missing evidence -> observation principle -> pose selection
- select the pose that best realizes the observation principle
- choose an appropriate move magnitude
- compose a spatial-clue narrative that bridges the new viewpoint back to the input views and to the question without stating the answer

**[Analysis]**
(*\ph{Question Analysis}*)

**[Question]**
(*\ph{question}*)

Now analyze the question and provide the JSON output.
\end{promptbox}
\newpage

\begin{promptbox}{Stage 2 - Final VQA}{prompt:s4}
You are answering a spatial question about a scene. You are given the original
input view(s) plus ONE additional synthesized view, rendered from a new camera
viewpoint to expose evidence that was hard to read in the originals.

**[Images]**
- Image 1-N: the original input view(s).
- Image N+1: a SYNTHESIZED view, produced by moving the camera
  ("(*\ph{magnitude}*)" "(*\ph{action}*)") starting from input Image (*\ph{reference\_view}*). It may
  contain rendering artifacts (*\textemdash{}*) treat it as supporting evidence, not ground truth,
  and cross-check it against the original views.

You previously analyzed this question and produced the following view planning:
(*\ph{Question Analysis}*)
(*\ph{Reasoning Guidance}*)

Use all the images together to answer the original question below.

**[Question]**
(*\ph{question}*)

(*\ph{benchmark answer-format instruction}*)
\end{promptbox}

\begin{promptbox}{Stage 2 - Fallback for Final VQA}{prompt:s4fb}
You previously analyzed this question and produced the following view planning:
(*\ph{Question Analysis}*)
(*\ph{Reasoning Guidance}*)

Now, using that analysis together with the input images, answer the original question.

**[Question]**
(*\ph{question}*)

(*\ph{benchmark answer-format instruction}*)
\end{promptbox}

\end{document}